\documentclass[preprint,12pt]{elsarticle}

\usepackage{lineno,graphicx,epsfig,setspace,subfig,url,amsmath}
\usepackage{amssymb}
\usepackage{algorithm}
\usepackage[noend]{algpseudocode}
\usepackage{longtable}
\usepackage{array}
\usepackage{mathtools}
\usepackage{caption}
\usepackage{multirow}
\usepackage{relsize}
\usepackage{color}
\usepackage{bm}
\modulolinenumbers[1]
\usepackage{lineno}
\journal{Journal of Medical Image Analysis}

\begin{document}

\begin{frontmatter}


\title{Pre and Post-hoc Diagnosis and Interpretation of Malignancy from Breast DCE-MRI}



\author[label0]{Gabriel Maicas \fnref{DP}\corref{correspondingauthor}}

\cortext[correspondingauthor]{Corresponding author}
\fntext[DP]{This work was partially supported by the Australian Research Council project (DP180103232).}
\ead{gabriel.maicas@adelaide.edu.au}

\author[label2]{Andrew P. Bradley\fnref{DP}}
\author[label1]{Jacinto C. Nascimento\fnref{DP}}
\author[label0]{\\Ian Reid\fnref{ARC}}
\author[label0]{Gustavo Carneiro\fnref{DP}}
\fntext[ARC]{IR acknowledges the Australian Research Council: ARC Centre for Robotic Vision (CE140100016) and Laureate Fellowship (FL130100102)}
\address[label0]{Australian Institute for Machine Learning, The University of Adelaide, Australia}
\address[label2]{Science and Engineering Faculty, Queensland University of Technology, Australia}
\address[label1]{Institute for Systems and Robotics, Instituto Superior Tecnico, Portugal}
\begin{abstract}
We propose a new method for breast cancer screening from DCE-MRI based on a post-hoc approach that is trained using weakly annotated data (i.e., labels are available only at the image level without any lesion delineation). Our proposed post-hoc method automatically diagnosis the whole volume and, for positive cases, it localizes the malignant lesions that led to such diagnosis.
Conversely, traditional approaches follow a pre-hoc approach that initially localises suspicious areas that are subsequently classified to establish the breast malignancy -- this approach is trained using strongly annotated data (i.e., it needs a delineation and classification of all lesions in an image).
Another goal of this paper is to establish the advantages and disadvantages of both approaches when applied to breast screening from DCE-MRI. 
Relying on experiments on a breast DCE-MRI dataset that contains scans of 117 patients, our results show that the post-hoc method is more accurate for diagnosing the whole volume per patient, achieving an AUC of 0.91, while the pre-hoc method achieves an AUC of 0.81.
However, the performance for localising the malignant lesions remains challenging for the post-hoc method due to the weakly labelled dataset employed during training.
\end{abstract}


\end{frontmatter}

%\linenumbers

%%%%%%%%%%%%%%%%%%%%%%%%%%%%%%%%%%%%%%%%%%%%%	Introduction
\section{Introduction}
\label{S:1}
Breast cancer is amongst the most diagnosed cancers~\citep{ausStats2017,cancerStats17} affecting women worldwide~\citep{desantis2015international,torre2015global}.
One of the most effective ways of increasing the survival rate for this disease is based on early detection~\citep{saadatmand2015influence,welch2016breast}.
Screening programs aim to provide such early detection by diagnosing at-risk, asymptomatic patients, allowing for an early intervention and treatment.
The most widely employed image modality for population-based breast screening is mammography. High risk patients are also recommended to undergo screening with dynamically contrast enhanced magnetic resonance imaging (DCE-MRI)~\citep{mainiero2017acr,smith2017cancer}. DCE-MRI is known to increase the sensitivity, compared to mammography, especially in young patients that have denser breasts~\citep	{kriege2004efficacy}.

However, the diagnosis and interpretation of DCE-MRI is a challenging and time consuming task that involves the interpretation of large amounts of data~\citep{behrens2007computer} and is prone to high inter-observer variability~\citep{grimm2015interobserver,lehman2013accuracy}.
Computer-aided diagnosis (CAD) systems are designed to reduce the analysis time~\citep{gubern2016automated,wood2005computer},  increase sensitivity~\citep{Vreemann2018} 
and specificity~\citep{meinel2007breast}, and serve as a second (automated) reader~\citep{shimauchi2011evaluation}. 
Designing such systems is challenging due to the variability in location, appearance~\citep{levman2009effect}, size and shape~\citep{song2016progress}, and the low signal-to-noise ratio~\citep{kousi2015quality} of lesions.
In general, such CAD systems can be categorised as pre-hoc or post-hoc, depending on how the processing stages are organised, as explained below.

\begin{figure}[t]
\begin{tabular}{ccc}
\subfloat[]{\includegraphics[width = 1.5in]{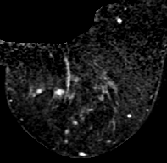}\label{img1}} &
\hspace{0.1in}
\subfloat[]{\includegraphics[width = 1.5in, height = 1.47in]{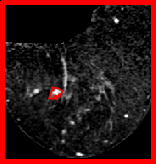}\label{img2}}&
\hspace{0.1in}
\subfloat[]{\includegraphics[width = 1.5in]{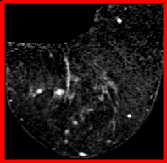}\label{img3}}
\end{tabular}
\caption{Example of a DCE-MRI breast image and annotation types.
Image (\ref{img1}) shows a slice of a breast DCE-MRI volume.
Image (\ref{img2}) shows the same slice with the strong annotations: lesion delineation classification as malignant.
Image (\ref{img3}) shows the weak annotation (i.e., whole image) of the same breast volume as malignant. }
\label{fig:preAndPost}
\end{figure}

Fully automated pre-hoc CAD methods for breast screening~\citep{amit2017hybrid,dalmics2018fully,gubern2015automated} from DCE-MRI compute the confidence score of malignancy of a breast using the following two-stage sequential approach: 1) detection of suspicious lesions, and 2) classification of the detected lesions.
During detection (i.e., first stage), the algorithm localises benign and malignant lesions, and possibly false positive detections, in the image, which are then classified as malignant or non-malignant in the second stage.  Four important challenges arise with this pre-hoc approach. 
Firstly, the modelling of the detector requires strong labels, i.e., precise voxel-wise annotation of lesions (see Fig.~\ref{fig:preAndPost} for an example of different types of annotations). Strong annotation is expensive because it requires experts to label a relatively large number of training volumes;  in addition, given the difficulties involved in such manual labelling process, this annotation may contain noise (this happens partly because experts are generally not trained to provide such precise annotations in regular practice).  
Secondly, the classifier may be trained using incorrect
manually annotated lesion class labels. Such manual annotation is usually produced by biopsy analysis, but if there are benign and malignant lesions jointly present in the same breast, this analysis may not determine the correct association. 
Thirdly, apart from rare exceptions that need large annotated training sets~\citep{ribli2018detecting}, pre-hoc diagnosis systems are generally trained in a two-stage process~\citep{gubern2015automated,mcclymont2015computer}. This pipeline is not the optimal way to maximise classification diagnosis performance because the final classification depends on the detection, but the detection optimality does not warrant classification optimality.
Finally, the fourth challenge is that the classification accuracy is limited by the detector performance, where it is impossible for the classifier to recover from a missing lesion detection because it can not be classified.

An alternative approach that is starting to gain traction~\citep{esteva2017dermatologist,maicas2018training,wang2017chestx} reverses these stages. The first stage aims to classify the whole breast scan directly, followed by a second stage that localizes regions in the scan that can explain the classification

-- for instance, if the first stage outputs a malignant diagnosis, then the second stage aims to find malignant lesions in the scan. We term this a {\em post-hoc} approach. %
This approach is of special interest for the problem of breast screening from DCE-MRI because the whole-scan diagnosis can, for example, analyse regions other than lesions that may contain relevant information for the diagnosis~\citep{kostopoulos2017computer}.
The main advantage of these systems compared to pre-hoc systems is the possibility of using scan-level labels (referred to as weak labels in the rest of the paper). Such labels are already present in many Picture Archiving and Communication Systems (PACS) or can be automatically extracted from radiology reports~\citep{wang2017chestx}, eliminating almost completely the effort needed for the manual annotation described above for the pre-hoc approach. 
Also, the use of scan-level labels overcomes the limitations in annotations required by pre-hoc approaches. Firstly, there is no need for lesion delineation avoiding such costly process. Secondly, the incorrect labelling of lesions explained above is reduced as the most likely lesion to be malignant is biopsied and therefore the label is more likely to be correct --there is no need to associate labels with lesions). 
The main challenge of post-hoc systems resides in highlighting the scan regions that can justify a particular classification (e.g., in the case of a malignant classification, it is expected that the regions represent the malignant areas of the scan), given that such manual annotation is not available.  This challenge is important for the deployment of post-hoc systems in clinical practice~\citep{caruana2015intelligible}.

\begin{figure}[h]
\centering\includegraphics[width=1.\linewidth]{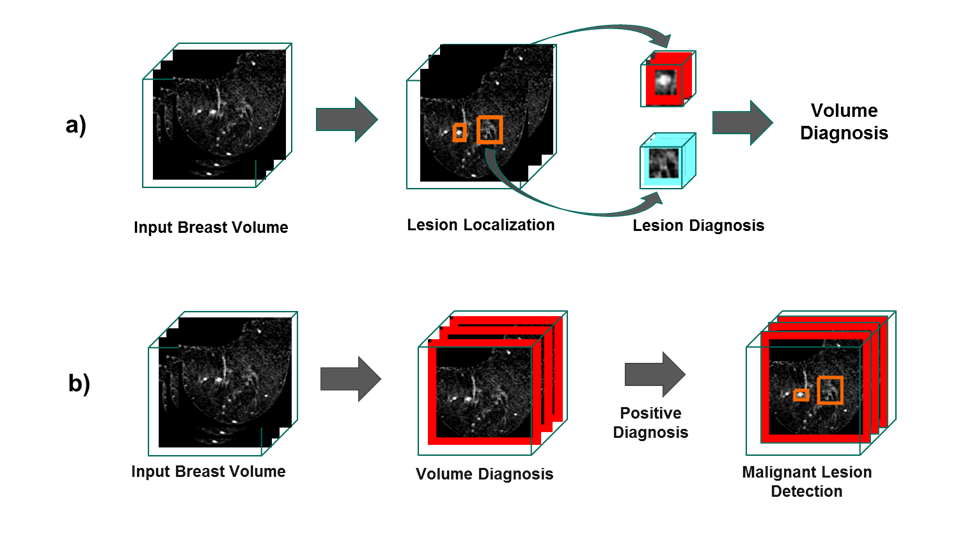}
\caption{Pre-hoc and post-hoc approaches for breast screening. \textbf{a)} The pre-hoc approach first localises lesions in the input breast volume (e.g., detections in orange), and then these lesions are classified to decide about their malignancy (e.g., red indicates positive and blue means negative malignancy classification). Finally, the breast volume is diagnosed according to the classification scores of the lesions. 
\textbf{b)} The post-hoc approach first diagnoses the input breast volume (e.g., red means positive malignancy classification). If the diagnosis is positive, then malignant lesions are localised in the breast (e.g., detections in orange).}
\label{fig:intro}
\end{figure}

In this paper, we propose a new post-hoc method and a
systematic comparison between pre-hoc and post-hoc approaches for breast screening from DCE-MRI. We aim to answer the following research questions: 1) which approach should be chosen if the goal is to optimally classify a whole scan in terms of malignant or non-malignant findings, and 2) how accurate is the localisation of malignant lesions produced by post-hoc approaches when compared with the localisation of malignant lesions produced by pre-hoc methods. 
The pre-hoc system considered in this paper is based on our recent detection model~\citep{maicas2017deep} that achieves state-of-the-art (SOTA) lesion localisation, while reducing the inference time needed by traditional exhaustive search methods.  
For the post-hoc system, we rely on our recently proposed approach based on meta-learning~\citep{maicas2018training} that holds the SOTA performance for the problem of breast screening from DCE-MRI. Decision interpretation is based on our recent 1-class saliency detector~\citep{maicas2018lesion}, especially designed for the weakly supervised lesion localisation problem after performing volume diagnosis.
See Fig.~\ref{fig:intro} for an overview of the pre-hoc and post-hoc pipelines.  

Experiments on a breast DCE-MRI dataset containing 117 patients and 141 lesions show that the post-hoc system achieves better malignancy classification accuracy than the pre-hoc method. In terms of lesion localisation, the post-hoc approach shows less accurate performance compared to the pre-hoc system, which we infer that is mostly due to the weak annotation used in the training phase of the post-hoc method.

%%%%%%%%%%%%%%%%%%%%%%%%%%%%%%%%%%%%%%%%%%%%%	Literature Review
\section{Literature Review}
\label{S:2}

\subsection{Pre-hoc Approaches}

Pre-hoc approaches are assumed to contain two sequential stages: 1) detection of regions of interest (ROI) containing suspicious tissue, and 2) classification of ROIs into malignant or not malignant (benign and/or false positive) tissue.

Traditional pre-hoc approaches for breast screening from breast DCE-MRI were based on manual~\citep{agner2014computerized,gallego2015improving,mus2017time,soares20133d} or semi-automated~\citep{chen2006fuzzy,dalmics2016computer,meinel2007breast,milenkovic2017textural,platel2014automated} ROI detection.
In addition, the classification in these traditional approaches was based on support vector machine (SVM), random forest, or artificial neural network models, using hand-designed features (e.g., dynamic, morphological, textural or multifractal)~\citep{dalmics2016computer,meinel2007breast,milenkovic2017textural,platel2014automated}.

Aiming at reducing user intervention to reduce the number of ROIs~\citep{liu2017total}, pre-hoc systems evolved to be fully automated. 
Such automated pre-hoc approaches generally employed an exhaustive search method 
or clustering to detect ROIs in the scan using hand-designed features~\citep{gubern2015automated,mcclymont2015computer,renz2012detection,wang2014robust}. The classification of ROIs into false positive, benign or malignant findings is then performed with a new set of hand-designed features extracted from the ROIs~\citep{gubern2015automated,mcclymont2015computer,renz2012detection,wang2014robust}.
These fully automated methods generally suffer from two issues: 1) the sub-optimality of hand-designed features needed at both ROI localization and ROI classification, and 2) the high computational cost of the exhaustive search to detect ROIs.

Both limitations have been addressed after the introduction of deep learning methodologies~\citep{krizhevsky2012imagenet} in the field of medical image analysis.
Initially, feature sub-optimality was addressed either for ROI detection~\citep{maicas2017globally,maicas2017deep} or classification~\citep{amit2017classification,amit2017hybrid,rasti2017breast}, 
but it was recently solved for both detection and classification~\citep{dalmics2018fully}. 
~\cite{dalmics2018fully} also reduced the inference time of the exhaustive search by directly computing a segmentation map from the scan using a U-net~\citep{ronneberger2015u}.

Although each step of the pipeline has been individually optimized, there is no guarantee that the full pipeline is optimal in terms of classification accuracy. This was addressed with the formation of large datasets that has enabled the use of SOTA one-stage detection and classification computer vision techniques, such as Faster R-CNN~\citep{ren2015faster} or Mask RCNN~\citep{he2017mask}. The main advantage of these methods lies in the optimality of the end-to-end training, effectively merging the detection and classification tasks~\citep{dalmics2018fully}. 
For example,~\cite{ribli2018detecting} applied Faster R-CNN to detect tumours from mammograms and they showed that this approach is quite efficient in terms of inference time. However, Faster R-CNN generalises poorly, which means that the training set must contain a large annotated set of ROIs and, at the same time, be rich enough to comprise all possible lesion  variations.  
Besides the need for large datasets, which are difficult to acquire for DCE-MRI breast screening,  these systems suffer from the need for strong annotations (i.e., the accurate delineation of the lesions). 
~\cite{li2018thoracic} partially addressed this issue by developing a semi-supervised system, alleviating the need of lesion annotations. However, a large number of annotated images (880) is still required to train the system.

\subsection{Post-hoc Approaches}

Post-hoc systems aim to overcome the need for strong annotations by training models with only scan-level labels (i.e., weak labels). This is especially useful for the problem of breast screening, where the analysis of adjacent regions to lesions may be important~\citep{kostopoulos2017computer}.  In addition, the classification accuracy of post-hoc systems are not constrained by the lesion detection, which is the case in pre-hoc systems.

Several post-hoc systems have been proposed~\citep{wang2017chestx,zhu2017deep}. 
For instance, 
~\cite{wang2017chestx} use a deep learning model to produce classification scores from whole scans and 
~\cite{zhu2017deep} propose a deep multiple instance learning.  
However, these approaches still require large datasets to achieve good performance.
This issue was addressed by 
~\cite{maicas2018training}, who proposed a new meta-learning methodology to learn from a small number of annotated training images. Their work established a new SOTA classification accuracy for breast screening from DCE-MRI.

The main challenge for post-hoc models arises from the fact that they do not use manually annotated ROIs for training, which makes the ROI localisation (and delineation) a hard task.  Such ROI localisation is important for explaining the classification made by the CAD system in clinical settings (e.g., for a scan classified as malignant, doctors are likely to know where the lesions are located). Solving this lesion localisation problem is a research problem that is being actively investigated in the field~\citep{dubost2017gp,feng2017discriminative,maicas2018lesion,wang2017zoom,yang2017joint}.
The approach proposed by 
~\cite{maicas2018lesion} achieves SOTA detection performance by properly defining saliency for the problem of weakly supervised lesion localisation, which assures that salient regions represent malignant lesions in the image.

However, the literature does not provide any studies comparing pre and post-hoc diagnosis approaches. The main reason for this absence of comparison among the methods described in this literature review is that such analysis is not straightforward due to~\citep{maicas2017deep}: 1) the lack of publicly available datasets that can be used to compare new approaches to the current state-of-the-art, 2) the criteria to decide if an ROI is a true positive detection, and 3) the criteria to decide if lesions labelled as the challenging BIRADS=3 should be included into the benign category~\citep{gubern2015automated}.
In addition, not all assessments of pre-hoc fully automated methodologies consider false positives in the diagnostic stage as they only differentiate between benign and malignant~\citep{mcclymont2015computer}. 
We propose to compare both types of automated approaches for the problem of breast screening from breast DCE-MRI. With the use of a common dataset and well-defined criteria to satisfy the issues described above, we investigate which approach performs better for breast diagnosis and lesion localisation. 

%%%%%%%%%%%%%%%%%%%%%%%%%%%%%%%%%%%%%%%%%%%%%	Methods
\section{Methods}
\label{S:3}
This section provides a formal description of the dataset in Sec. \ref{sec:dataset}, the pre-hoc method in Sec. \ref{sec:pre-hoc}, and the post-hoc approach in Sec. \ref{sec:pos-hoc}.

%%%%%%%%%%%%%%%%%%%%%%%%%%%%%%%%%%%%%%%%%%%%%	Dataset
\subsection{Dataset}
\label{sec:dataset}

Let {\scriptsize ${\cal D} = \left \{ \left ( \mathbf{b}_i, \mathbf{x}_i, \mathbf{t}_i, \{ \mathbf{s}_i^{(j)} \}_{j = 1}^M, \{ \mathbf{l}_i^{(j)} \}_{j = 1}^M, \mathbf{y}_i\right )_i \right \}_{i \in \{1,...,|{\cal D}| \},\mathbf{b}_i \in \{ \text{left},\text{right} \}}$} denote the 3D DCE-MRI dataset, where
$\mathbf{b}_i \in \{ \text{left},\text{right} \}$ specifies the left or right breast of the $i^{th}$ patient; 
$\mathbf{x}_i,\mathbf{t}_i:\Omega \rightarrow \mathbb R$ represent the first 3D DCE-MRI subtraction volume and the T1-weighted MRI volume used for preprocessing, respectively, with $\Omega\in\mathbb {R}^3$ representing the volume lattice of size $w\times h\times d$;
$\mathbf{s}_i^{(j)}:\Omega \rightarrow \{0,1\}$ is the voxelwise annotation of the $j^{th}$ lesion present in the breast $\mathbf{b}_i$ ($\mathbf{s}_i^{(j)}(\omega)=1$ indicates the presence of lesion in voxel $\omega \in \Omega$, and $\mathbf{s}_i^{(j)}(\omega)=0$ denotes the absence of lesion);
$\{ \mathbf{l}_i^{(j)} \}_{j = 1}^M \in \{ 0,1 \}$ indicates the classification of lesion $j$ as benign or malignant, respectively; 
and  $\mathbf{y}_i$ is a scan-level label with the following values: $\mathbf{y}_i = 0$ if there is no lesion in breast $\mathbf{b}_i$, $\mathbf{y}_i = 1$ if all the lesion(s) in breast $\mathbf{b}_i$ are benign or $\mathbf{y}_i = 2$ if there is at least one malignant lesion.
The dataset is patient-wise split into train $\mathcal{T}$, validation $\mathcal{V}$ and test $\mathcal{U}$ sets, such that images of each patient only belong to one of the sets.
Note that the voxelwise lesion annotations $\{ \mathbf{s}_i^{(j)} \}_{j = 1}^M \text{ and } \{ \mathbf{l}_i^{(j)} \}_{j = 1}^M$ are not employed during the training of the post-hoc system -- they are only used to train and test the pre-hoc system and in the quantification of the results for both systems.
Finally, the motivation behind the use of the first subtraction image $\mathbf{x}$ lies in the reduction of cost and time for image acquisition and analysis~\citep{gilbert2018personalised,mango2015abbreviated}.

%%%%%%%%%%%%%%%%%%%%%%%%%%%%%%%%%%%%%%%%%%%%%	Model
\subsection{Pre-hoc Method}\label{sec:pre-hoc}

Our proposed pre-hoc approach is based on the following steps:
\begin{enumerate}
\item \textbf{Lesion detection} (Sec.~\ref{sec:pre-hoc-training-inference}): an attention mechanism based on deep reinforcement learning (DRL)~\citep{mnih2015human} searches for lesions using a method that analyses large portions of the breast volume and iteratively focuses the search on the appropriate regions of the input volume. 
\item \textbf{Lesion diagnosis} (Sec.~\ref{sec:classification}): a state-of-the-art deep learning classifier~\citep{huang2017densely} analyses the lesions detected in the previous step in order to classify them as malignant or non-malignant (note that non-malignant regions are represented by benign lesions or normal tissue, i.e. false positive detections). 
The confidence score of malignancy for the breast volume is defined as the maximum probability of malignancy among the detected lesions. 
\end{enumerate}

\subsubsection{Lesion Detection}\label{sec:pre-hoc-training-inference}

\begin{figure}[h]
\centering\includegraphics[width=1.\linewidth]{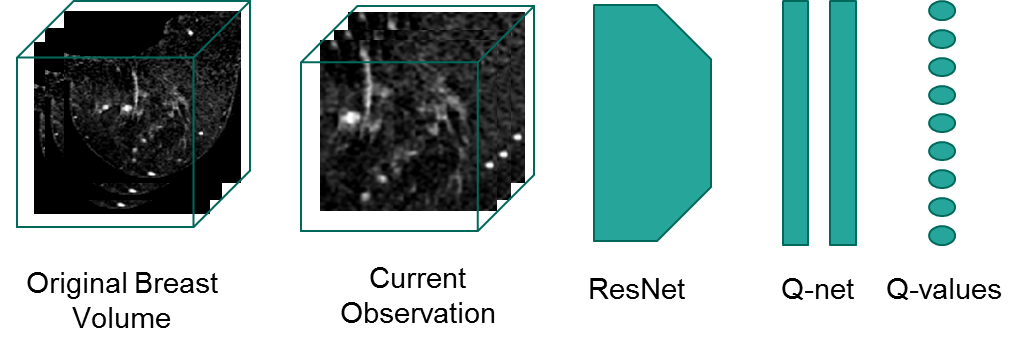}
\caption{Overview of the proposed lesion detection method. The bounding volume of the current observation is extracted from the input breast volume and fed to the 3D ResNet to obtain the embedding of the observation. The embedding is then forwarded through the Q-net to obtain the Q-values for each of the actions.}
\label{fig:RLprocess}
\end{figure}

We propose an attention model that is capable of reducing the inference time of previous methods for lesion detection~\citep{gubern2015automated,mcclymont2014fully} in pre-hoc systems.
This attention mechanism searches for lesions by progressively transforming relatively large initial bounding volumes (BV) ({\em i.e.} sub-regions of the {\em MRI volume}) into smaller regions containing a more focused view of potential lesions~\citep{maicas2017deep}. 
The transformation process is guided by a policy $\pi$ that indicates how to optimally change the current BV to detect a lesion. The policy is represented by a deep neural network, called deep Q-net (DQN), that receives as input an embedding vector $\mathbf{o} \in \mathbb R^{O}$ of the current BV and outputs a measurement (i.e., the Q-value $(Q)$), representing the optimality associated with each of the possible transformations to find a lesion. See Figure~\ref{fig:RLprocess} for a block diagram of this process.
The aim of the learning phase is to model such policy, i.e., find the optimal parameters of the DQN. The inference exploits the policy to detect the lesions present in a breast DCE-MRI volume.

The training process of the DQN follows that of a traditional Markov Decision Process (MDP), which models a sequence of decisions to accomplish a goal from an initial state. At every time step, the current BV, represented by the observations $\mathbf{o}$, will be transformed by an action $a$, yielding a reward $r$ -- this reward indicates the effectiveness of the the chosen transformation for detecting a lesion.
The goal is to learn what actions should be applied to transform the current observation to another one with larger Dice coefficient measured with respect to the target lesion. In an MDP set-up, this translates into choosing the action that maximizes the expected sum of discounted future rewards~\citep{mnih2015human}: $R_t = \sum_{t'=t}^T \gamma^{t'-t} r_{t'}$, where $\gamma \in (0,1)$ is a discount factor.

Let $Q^{\star}(\mathbf{o},a)$ be the optimal Action-Value Function representing the expected sum of discounted future rewards by choosing action $a$ to transform the observation $\mathbf{o}$.  The optimal Action-Value function follows the policy $\pi$, as in:
\begin{equation}
Q^{\star}(\mathbf{o},a) = \max_{\pi} \mathbb{E}[R_t|\mathbf{o}_t=\mathbf{o},a_t=a,\pi].
\label{eq:training_Q1}
\end{equation}
Intuitively, $Q^{\star}(.)$ represents the {\em quality} of performing the action $a$ given the current observation $\mathbf{o}$ to achieve the final goal.  Therefore, the goal of the training process is to learn $Q^{\star}(.)$, which  maximizes the commulative sum of expected discounted rewards.

The optimal $Q^{\star}(\mathbf{o},a)$ can be computed iteratively using the {\em Bellman} equation and the Q-Learning algorithm ~\citep{Sut1998}: 
\begin{equation}
Q_{i+1}(\mathbf{o}_t,a_t) = \displaystyle{\mathop{\mathbb{E}}}_{\mathbf{o}_{t+1}}\Bigl[r_t+\gamma \max_{a_{t+1}} Q(\mathbf{o}_{t+1},a_{t+1})|\mathbf{o}_t,a_t\Bigr].\label{eq:bellman}    
\end{equation}

However, since it is impractical to compute $Q(\mathbf{o}_t,a_t)$ due to the large size of the observation-action space, a DQN function approximator, represented by $Q(\mathbf{o},a,\bm{\theta})$, can be used.
The weights $\bm{\theta}$ of the DQN $Q(\mathbf{o}_t,a_t,\bm{\theta}_t)$ can be learned by minimizing the mean square error of the Bellman equation:
{\small\begin{equation}
L(\bm{\theta}_t) = \mathbb{E}_{(\mathbf{o}_t,a_t,r_t,\mathbf{o}_{t+1})~\sim~U(\mathcal{E})} \Bigl[\Bigl(\underbrace{r_t + \gamma \max_{a_{t+1}}Q(\mathbf{o}_{t+1},a_{t+1};\bm{\theta}_t^-)}_{\rm target} - Q(\mathbf{o}_t,a_t;\bm{\theta}_t) \Bigr)^2 \Bigr],
\label{eq:Loss}
\end{equation}}where $\bm{\theta}_t$ are the parameters of the DQN at iteration $t$, $\bm{\theta}_t^-$ are the weights of the target network (defined below) used to compute the target value at iteration $t$, 
and $U(\mathcal{E})$ is a batch of experiences uniformly sampled from the experience replay memory $\mathbf{\mathcal{E}}_t$ (also defined below). 
The target network is used to compute the target values for each update of the weights of the DQN.
The architecture of this target network is the same as that of the DQN and its parameters $\bm{\theta}_t^{-}$ contain the weights of the DQN at a previous iteration of the optimization process. The weights $\bm{\theta}_t^{-}$ are updated after every iteration through the entire training set from the parameters $\bm{\theta}_t$ at the iteration $t-1$ and maintained constant between updates: $\bm{\theta}_t^- = \bm{\theta}_{t-1}$.
The experience-replay memory $\mathcal{E}_t=\{e_1,...,e_t\}$  stores previous experiences denoted by $e_t=\{\mathbf{o}_t,a_t,r_t, \mathbf{o}_{t+1} \}$, where each $e_t$ is collected at time step $t$ by choosing the action $a_t$ to transform from $\mathbf{o}_t$ into $\mathbf{o}_{t+1}$, yielding the reward $r_t$.
We describe in the next paragraphs how to obtain the observations, to choose the actions and to compute the reward function.

The embedding $\mathbf{o}$ of the current BV is computed as:
\begin{equation}
\mathbf{o} = f_{ResNet}({\bf x}(\bf b) , \mathbf{\theta_{ResNet}} )\label{eq:BB}
\end{equation}
where ${\bf b} = [b_x, b_y,b_z,b_w,b_h,b_d]\in\mathbb{R}^{6}$ is a bounding volume, with the triplets $(b_x, b_y,b_z)$ and $(b_w,b_h,b_d)$ denoting the top-left-front and the lower-right-back corners of the bounding volume, respectively; the DCE-MRI 
data is represented by ${\bf x}$; and $f_{ResNet}(.)$ represents a 3D Residual Network (ResNet)~\citep{he2016deep}.
The training of the 3D ResNet in \eqref{eq:BB} relies on a binary loss function that differentiates between input bounding volumes with and without lesions.
The dataset to train this 3D ResNet is built by sampling random BVs that are labelled as positive if the Dice Coefficient with a ground truth lesion is larger than 0.6, and negative otherwise. 
Note that the training of the 3D ResNet with a potentially infinite number of BVs from different scales, sizes and locations allows us to obtain a rich collection of BVs without the need for a large training set.

The set ${\cal A}=\{l_{x}^{+},l_{x}^{-},l_{y}^{+},l_{y}^{-},l_{z}^{+},l_{z}^{-},s^{+},s^{-},w \}$ represents the actions to modify the current BV, where $\{l,s,w\}$ represent the translation, scale and trigger (to terminate the search for lesions) actions, respectively; 
the subscripts $\{x,y,z\}$  denote the horizontal, vertical and depth translation, and the superscripts $\{+,-\}$ represent the positive/negative translation or up/down scaling. 

The reward function depends on the improvement in the lesion localisation process after selecting a specific action. 
For action $a \in {\cal A}\setminus \{w\}$, we measure the improvement in terms of the variation of the {\em Dice coefficient} after applying action $a$ to transform the observation $\mathbf{o}_{t}$ to $\mathbf{o}_{t+1}$:
\begin{equation}
r(\mathbf{o}_t,a,\mathbf{o}_{t+1}) = sign( d(\mathbf{o}_{t+1},\mathbf{s}) - d(\mathbf{o}_t,\mathbf{s})),
\label{eq:reward}
\end{equation}        
where $d(.)$ is the Dice coefficient between the bounding volume $\mathbf{o}$  and the ground truth $\mathbf{s}$.
The intuition behind (\ref{eq:reward}) is that the reward is positive if the Dice coefficient from observation $\mathbf{o}_t$ to observation $\mathbf{o}_{t+1}$ increases, and the reward is negative otherwise.
The quantization in (\ref{eq:reward}) avoids a deterioration of the training convergence due to small changes in $d(.)$~\citep{caicedo2015active}.

The reward for the trigger action, $a=w$, is defined as:
\begin{equation}
r(\mathbf{o}_t,a,\mathbf{o}_{t+1}) = 
\begin{cases}+\eta  & {\rm if}\;\; d(\mathbf{o}_{t+1},\mathbf{s}) \geq \tau_w\\ 
		-\eta & {\rm otherwise}
\end{cases}\label{eq:reward_trigger}
\end{equation}        
where $\eta > 1$ encourages the trigger action to finalize the search for lesions if the Dice coefficient with the ground truth $\mathbf{s}$ is larger than a pre-defined threshold $\tau_w$.

Actions during the training process are selected according to a modified $\epsilon$-greedy strategy to balance {\em exploration} and {\em exploitation}~\citep{maicas2017deep}:  with probability $\epsilon$, a random  action will be explored, and with probability $1-\epsilon$, the action will be chosen from the current policy.
During {\em exploration}, 
with probability $\kappa$, a random action is selected, and with probability $1 - \kappa$, a random action from the actions that will produce a positive reward is selected.
During  {\em exploitation}, 
the action is selected according to the current policy: $a_t=\arg\max_{a_t}Q(\mathbf{o}_t,a_t; \bm{\theta}_t)$.
The training process starts with $\epsilon=1$, which decreases linearly, transitioning from pure exploration to mostly exploitation following the current policy as the model learns to detect lesions.

During \textbf{inference}, 
we exploit the learned policy to detect lesions. In practice, we propose several initial bounding volumes covering different relatively large portions of the DCE-MRI volume.
Each initialization is processed independently and is iteratively transformed according to the action $a^{\star}_t$ indicated by the optimal action-value function:
\begin{equation}
a^{\star}_t = \arg\max_{a_t} Q(\mathbf{o}_t,a_t; \bm{\theta}^{\star}).
\end{equation}
where $\bm{\theta}^{\star}$ represents the parameter vector of the trained DQN model learned with \eqref{eq:Loss}.

We define the set of detected lesions as $\mathcal{D}^{pre} = \{ \mathcal{D}_i^{pre}\}_{i=1}^{|\mathcal{D}^{pre}|}$, 
where $\mathcal{D}_i^{pre}$ represents the $i^{th}$ bounding volume, when the trigger action is selected to stop the inference process. If the trigger action is not selected after 20 iterations, the search for a lesion is stopped yielding no detection.

\subsubsection{Lesion diagnosis}\label{sec:classification}

The detected lesions in $\mathcal{D}^{pre}$, formed during the lesion localization stage, are classified in terms of their malignancy. 
This binary classification is performed with a 3D DenseNet~\citep{huang2017densely}, trained using the detections from the training set to differentiate normal tissue and benign lesions (i.e., negative diagnoses) from malignant lesions (positive diagnosis).
During inference, each detection $\mathcal{D}_i^{pre}$ is fed through the 3D DenseNet to obtain its probability of malignancy.
Finally, the confidence score of malignancy of a breast  is defined as the maximum of the  malignancy probabilities computed from all the detected regions in such breast.
The confidence score of malignancy for the breast volume with no detections is set to zero.

\subsection{Post-hoc Method}\label{sec:pos-hoc}

Our proposed post-hoc approach is characterised by the following steps:
\begin{enumerate}
\item \textbf{Diagnosis} (Sec.~\ref{sec:VolDiagnose}): the classifier outputs the probability that a breast DCE-MRI volume contains a malignant lesion. Given the small training dataset, the model is first meta-trained with a teacher-student curriculum learning strategy to learn to solve several tasks. Then, the classifier is fine-tuned to solve the breast screening diagnosis task.
\item \textbf{Lesion Localization} (Sec.~\ref{sec:VolInterpretation}): the detector is weakly-trained to localise malignant lesions on breast DCE-MRI volumes that have been positively classified in the diagnosis stage above.  This lesion localisation process can be used to interpret the decision from the diagnosis stage. 
\end{enumerate}

\subsubsection{Breast Volume Diagnosis}
\label{sec:VolDiagnose}

Meta-training aims to learn a model that can solve new given tasks (classification problems) as opposed to traditional classifiers that solve a specific classification problem. Traditionally, models for solving new tasks have been achieved by fine-tuning pre-trained models~\citep{tajbakhsh2016convolutional}. However, these pre-trained models are rarely available for 3D volumes and large datasets are still required. These limitations can be overcome by including a meta-training phase before training, where the model is presented with several classification tasks that need to be solved, where each task has a small training set. Eventually, the model learns to solve new tasks that contain small training sets.

As noted in our previous work~\citep{maicas2018training}, the order in which to present classification tasks during meta-training influences the ability of the model to solve new tasks. Therefore, we propose to use the teacher-student curriculum learning strategy~\citep{matiisen2017teacher} that has been shown to outperform other strategies~\citep{maicas2018training}.

\begin{figure}[h]
\centering\includegraphics[width=1.\linewidth]{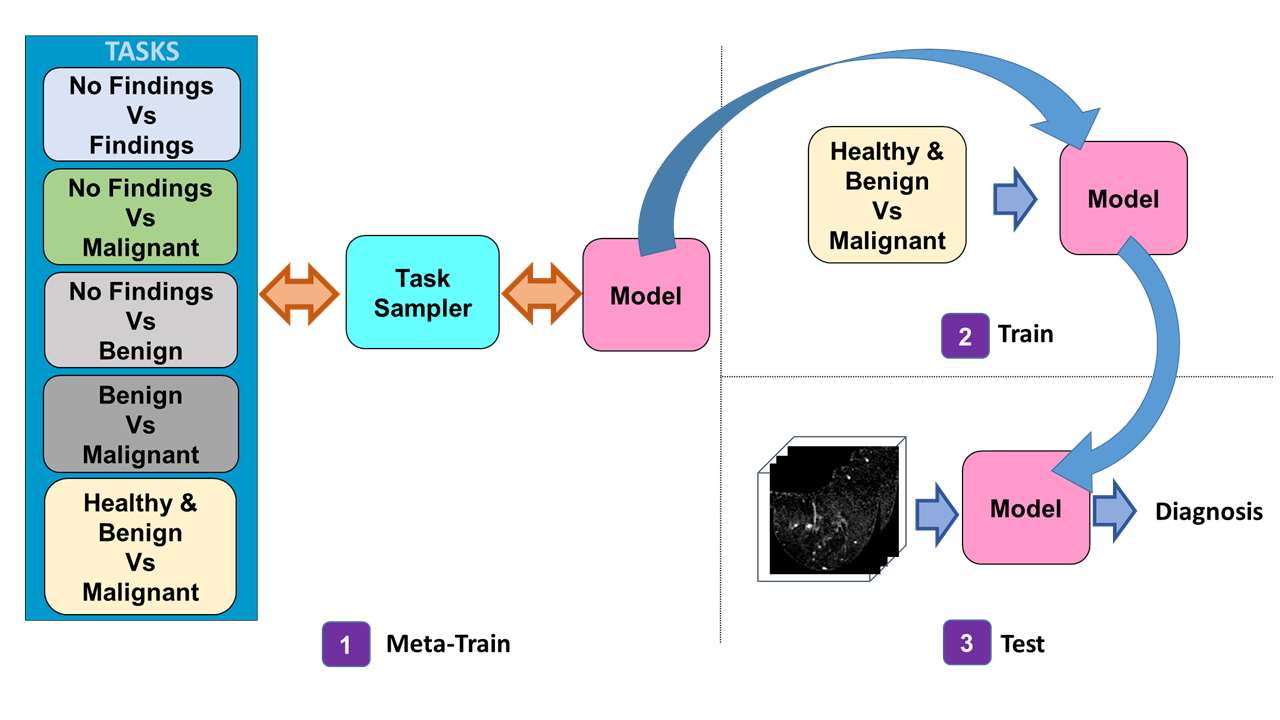}
\caption{Volume diagnosis process. Firstly, the model is meta-trained on several related classification tasks. Secondly, the model is trained in the breast screening task. Finally, the model is tested on the breast screening task.}
\label{fig:mlprocess}
\end{figure}

We propose to  meta-train the model to solve several related classification tasks, each containing a relatively small number of training images instead of training a classifier to distinguish volumes with any malignant findings from others containing no malignant lesions.
Firstly, during the meta-training phase, our model learns to solve different tasks that are formed from our breast DCE-MRI datasets. The tasks to be presented to the model are selected via the teacher-student curriculum learning strategy and contain a small training set. 
Secondly, the training phase is similar to that of any traditional classifier and solves the breast screening task using the samples available from the training set. The difference in our approach lies in the employment of the meta-trained model as the initialization for the training process.
As a result, when the meta-trained model is fine-tuned on the breast screening task with the small training set, it is able to efficiently and effectively classify previously unseen volumes containing malignant findings~\citep{maicas2018training}.
Finally, the inference phase (or breast diagnosis) consists of feeding the input volumes to the classifier to estimate the probability that they contain a malignant finding.
See Figure~\ref{fig:mlprocess} for an overview of the volume diagnosis process.

During \textbf{meta-training},  the model is meta-trained to solve the following five classification tasks: 
\begin{enumerate}
\item $K_1:$ findings (lesions) versus no findings,
\item $K_2:$ malignant findings versus no findings,
\item $K_3:$ benign findings versus no findings,
\item $K_4:$ benign findings versus malignant findings,
\item $K_5:$ malignant findings versus no malignant findings (i.e., breast screening). 
\end{enumerate}

Let $K= \cup_{i=1}^5  K_i $, where each task $K_i$ is associated with a dataset $\mathcal{D}_i$ that contains the volumes from the training set that are relevant for the task $K_i$. We define the meta-training set $\mathcal{D} = \cup_{i=1}^5  \mathcal{D}_i $.

Let the model to be meta-trained be defined by $g_{\theta}$ and the meta-update step be indexed by $t$. For each meta-update, a meta-batch $\mathcal{K}_t$ of tasks is sampled and contains $|\mathcal{K}_{t}|$ tasks from $K$ (see bellow for a description of the task sampling method). For each of the tasks $K_j \in \mathcal{K}_t$,  $N=N^{tr} + N^{val}$ volume-label samples are sampled from the corresponding meta-training set $\mathcal{D}_j$ to form $\mathcal{D}_j^{tr}$. 
Let $\mathcal{D}_j^{tr}$ contain $N^{tr}$ samples that will be used as training set and $\mathcal{D}_j^{val}$ contain $N^{val}$ samples that will be used as validation set during the $t^{th}$ meta-update for the $j^{th}$ task.

For every task $K_j \in \mathcal{K}_t$ in the meta-batch, the model is trained with $\mathcal{D}_j^{tr}$ to adapt to the task by performing several gradient descent updates. For simplicity, the adaptation of the model with one gradient descent update is defined by:
\begin{equation}
\theta_{j}^{\prime(t)} = \theta^{(t)} - 
\alpha \frac{\partial \mathcal{L}_{K_j} \left( g_{\theta^{(t)}} \left ( \mathcal{D}_j^{tr} \right) \right) }{\partial \theta},
\label{eq:innerUpdt}
\end{equation}
where $\theta^{(t)}$ are the parameters of the model at meta-iteration $t$, 
$\mathcal{L}_{K_j} (g_{\theta^{(t)}} \left ( \mathcal{D}_j^{tr} \right)) $ is the cross-entropy loss computed from $\mathcal{D}_j^{tr}$ for task ${K_j}$,
$\alpha$ is the learning rate for model adaptation, and
$\theta_{j}^{\prime(t)}$ are the adapted parameters after performing model adaptation for task ${K_j}$.

The adapted models $g_{\theta_{j}^{\prime(t)}}$ are subsequently evaluated with the validation pairs $\mathcal{D}_j^{val}$ of the corresponding task. The loss produced by the validation set on each of the tasks is used to compute the meta-gradient associated to each task. Finally, the model parameters $\theta$ are updated using the average of the meta-gradients associated to each of the tasks in the meta-batch:
\begin{equation}
\theta^{(t+1)} = \theta^{(t)} - \beta  \mathlarger{\sum_{K_j \in \mathcal{K}_m}} \frac{ \partial \mathcal{L}_{K_j} \left( 
g_{\theta_j^{\prime(t)}}\left( \mathcal{D}_j^{val} \right)\right )}{\partial \theta},
\label{eq:metaUpdt}
\end{equation}
where $\beta$ is the meta-learning rate and 
$\mathcal{L}_{K_j} \left( g_{\theta_j^{\prime(t)}}\left( \mathcal{D}_j^{val} \right)\right )$ is the cross entropy loss of the validation volumes in $\mathcal{D}_j^{val}$ for task ${K_j}$.
This procedure is repeated for  $M$ meta-iterations, as shown in Alg.~\ref{alg:meta-training}.  
\begin{algorithm}
\caption{Overview of the meta-training procedure presented in~\citep{maicas2018training}}\label{alg:metatraining}
\begin{algorithmic}[]
\footnotesize{
\Procedure{Meta-train}{$\{K_1 \ldots K_5\}$, $\{\mathcal{D}_1 \ldots \mathcal{D}_5\}$, model $g_{\theta}$}
\State Initialise parameters $\theta$ from  $g_{\theta}$
	\For{$t=1\;\; to \;\; T$} 
    \State \textbf{Sample} meta-batch $\mathcal{K}_{t}$ by sampling $|\mathcal{K}_{t}|$ tasks from $\{K_1 \ldots K_5\}$ 
        \For{each task $K_j \in \text{meta-batch }\mathcal{K}_t$}
		\State \textbf{Adapt} model using (\ref{eq:innerUpdt}) with samples from $\mathcal{D}_j^{tr}$ 
        \State \textbf{Evaluate} adapted model using with samples from $\mathcal{D}_j^{val}$ 
		\EndFor\label{endEveryTask}
	\State \textbf{Meta-update} model parameters with (\ref{eq:metaUpdt})
	\EndFor\label{endMetaUpdate}
\EndProcedure
}
\end{algorithmic}
\label{alg:meta-training}
\end{algorithm}

The breast screening \textbf{training} process is initialised by the meta-trained model.  Using the entire training set  $\mathcal{T}$, the model adapts to the breast screening task by performing several gradient descent updates, similarly to the training of a traditional deep learning classifier. We use the validation set $\mathcal{V}$ for model selection.
The \textbf{inference} of the model is similar to that of any standard classifier and consists of feeding the testing volume through the network to obtain the probability of malignancy 
of each of the input volumes. The confidence score of malignancy corresponds to the probability of malignant output by the classifier.

During the \textbf{meta-learning process}, the \textbf{task sampling} process to form a meta-batch of tasks depends on the past observed performance improvements of the model in each of the tasks. This has been shown to outperform other alternative approaches~\citep{maicas2018training}. % 
A partially observable Markov decision process (POMDP) solved using reinforcement learning with Thompson Sampling can model such an approach. A POMDP is characterized by observations, actions, and rewards. In our set-up, we define an observation $O_{K_j}$ 
as the variation in the area under the receiving operating characteristic curve (AUC) of the adapted model $\theta_{j}^{\prime(t)}$ compared to the initial AUC before the model  $\theta^{(t)}$  was adapted to the task $K_j \in \mathcal{K}_t$ -- in both cases, the AUC is measured using the sampled validation set $\mathcal{D}_j^{val}$. The actions correspond to sampling a particular task. The reward is defined as the difference between the current and previous observations during the last time that the task was sampled.
The goal is to decide which action to apply, i.e. which task should be sampled for the next meta-training iteration. We use Thompson sampling to decide the next task to be sampled, which allows us to balance between sampling new tasks, and sampling tasks for which the improvement of performance is currently higher (similar to the exploration-exploitation dilemma in reinforcement learning)~\citep{matiisen2017teacher}.

Let $\mathcal{B}_j$ be a buffer of recent rewards for task $K_j$ -- this buffer stores the last $B$ rewards for this task.
To perform Thompson sampling, a random recent reward $R_{\mathcal{B}_j} \in \mathcal{B}_j$ is uniformly sampled. The next task $K_j$ to be included in the meta-batch $\mathcal{K}_{t}$ of iteration $t$ is selected with $j = \arg\max_i |R_{\mathcal{B}_i}|$. This process is repeated for $|\mathcal{K}_{t}|$ times to form $\mathcal{K}_{t}$.
The intuition behind this is that for tasks where performance is increasing rapidly (i.e. yielding higher rewards) they will be sampled more frequently until mastered (i.e. the reward will tend to zero as the variation in AUC after adaptation will tend to be smaller in consecutive iterations). Then, a different task will be sampled more frequently. However, if the model reduces the performance in the previously mastered task, it will be sampled again more frequently because the absolute value of the reward will tend to be higher again.

\subsubsection{Malignant Region Localization}
\label{sec:VolInterpretation}

A breast volume is diagnosed as malignant in the previous step if its confidence score of malignancy is higher than the equal error rate (EER) of the proposed classifier on the validation set. %  
The EER as threshold is chosen to avoid any preference between sensitivity and specificity. % 
For positively classified volumes, we aim to generate a saliency map represented by a binary mask indicating the localization of lesions that can explain the decision made by the classifier; while for negatively 
classified volumes, no salient region is produced. Therefore, we propose a 1-class saliency detector~\citep{maicas2018lesion} that has been specifically designed to satisfy these conditions.

Our 1-class saliency detector is modelled with a weakly-supervised training process to detect salient regions in positively classified volumes, where these regions denote malignant lesions. The detector follows an encoder-decoder architecture that generates a mask $\mathbf{m}:\Omega \rightarrow [0,1]$ of the same size as the input volume, where this mask localizes the most salient regions of the input volume that are involved in the positive classification. 
The encoder is the classifier from Sec.~\ref{sec:VolDiagnose}, which produces the diagnosis. The decoder up-samples the output from the encoder to the original resolution from the lowest resolution feature maps by concatenating four blocks of feature map resize, convolution layer, batch normalization layer and ReLU activation~\citep{zeiler2014visualizing}. Skip connections are used to connect corresponding layers of the same resolution in the encoder and decoder.
During training, the parameters of the encoder are fixed and the parameters of the decoder are updated using the gradient corresponding to the following loss for each volume $\mathbf{x}_i$:
\begin{equation}
\ell_i(\mathbf{m})= \lambda_1 \ell_{TV}(\mathbf{m}) + \lambda_2 \ell_{A}(\mathbf{m}) - y_i \lambda_3  \ell_{P}(\mathbf{m},\mathbf{x}_i) + y_i \lambda_4 \ell_{D}(1-\mathbf{m},\mathbf{x}_i),
\label{eq:PostweakLocLoss}
\end{equation}
where
$\ell_{TV}$ measures the total variation of the mask forcing the boundary of salient regions to be relatively smooth,
$\ell_{A}$ measures the area of the salient regions and aims to reduce the total area of regions,
$\ell_{P}$ measures the confidence in the classification of the input volume  $\mathbf{x}_i$ masked with $\mathbf{m}$, and
$\ell_{D}$ measures the confidence in the classification of the input volume  $\mathbf{x}_i$ masked with the inverse of the generated mask, i.e $(1 - \mathbf{m} )$. %  

By \textbf{training} the mask generator model with the loss function~\eqref{eq:PostweakLocLoss}, there is an explicit relationship between saliency and malignant lesions~\citep{maicas2018lesion}. By setting $y_i=0$ for negative volumes, they are forced to have no salient regions. For positives volumes, salient regions are forced to have the following characteristics: 1) be small and smooth, 2) when used to mask the input volume, the classification result is positive; and 3) when its inverse is used to mask the input volume, the classification result is negative. 
During \textbf{inference}, volumes diagnosed as positive are fed forward through the decoder to produce a mask, where each voxel has values in $[0,1]$. This mask is thresholded at $\zeta$ to obtain the malignant lesions.

%%%%%%%%%%%%%%%%%%%%%%%%%%%%%%%%%%%%%%%%%%%%%	Experiments
\section{Experiments}
\label{S:4}

In this section, we describe the dataset and experimental set-up used to assess the proposed methods for the problems of breast screening and malignant lesion detection.

%%%%%%%%%%%%%%%%%%%%%%%%%%%%%%%%%%%%%%%%%%%%%	Dataset - Experiments
\subsection{Dataset}

Our methods are evaluated with a dataset containing MRI scans from 117 patients. The dataset is patient-wise split into training, validation and test sets using the same split as previous approaches~\citep{maicas2017deep,maicas2018training,maicas2018lesion}. The training set contains scans from 45 patients, where these scans show 38 malignant lesions and 19 benign lesions -- the scans also show that 29 of the patients have at least one malignant lesion while 16 only have benign lesion(s). 
The validation set has scans from 13 patients, with 11 malignant and 4 benign lesions 
-- these scans show that 9 of the patients have at least one malignant lesion while 4 patients have only benign lesion(s). The test set contains scans from 59 patients, with  46 malignant and 23 benign lesions -- the scans show that 37 of the patients have at least one malignant lesion while 22 have only benign lesion(s). 
The characterization of each lesion was confirmed with a biopsy. 
Every patient has at least one lesion, but not every breast contains lesions. 
There are 42, 13, and 58 breasts with no lesions in the training, validation and testing sets, respectively. Likewise, 18, 4, and 22 breasts contain only benign lesions (i.e. are considered ``benign'') and 30, 9, and 38 contain at least one malignant lesions (i.e. are considered ``malignant'').
For the breast screening problem, ``Malignant'' breasts are considered positive while ``benign'' and breasts with no lesions are considered negative.  
The MRI dataset~\citep{mcclymont2014fully} contains T1-weighted and two dynamic contrast enhanced (pre-contrast and first post-contrast) volumes for each patient acquired with a 1.5 Tesla GE Signa HDxt scanner.
The   T1-weighted   anatomical   volumes  were acquired  without  fat  suppression and with an acquisition matrix of $512 \times 512$.
The DCE-MRI images are based on T1-weighted volumes with fat suppression, with an  acquisition matrix of $360 \times 360$ and a slice thickness of 1 mm. Firstly, a pre-contrast volume was acquired before a contrast agent was injected. The first post-contrast volume was acquired after a delay of 45 seconds after the acquisition of the pre-contrast. The first subtraction volume is formed by subtracting the pre-contrast volume to the first post-contrast volume. 
Both T1-weighted and DCE-MRI were acquired axially.

The dataset was preprocessed using the T1-weighted volume to segment the breast region from the chest wall using Hayton's method~\citep{hayton1997analysis,mcclymont2014fully}. This involves removing the pectoral muscle which may produce false positive detections.  In addition, the breast region was divided into left and right breasts by splitting the volume in halves, as the breast region was initially centred.
Each breast volume was resized to a size of $100\times 100 \times 50$ voxels. Note that we operate the proposed methods breast-wise.  

%%%%%%%%%%%%%%%%%%%%%%%%%%%%%%%%%%%%%%%%%%%%%	Experimental set-up
\subsection{Experimental Set-up}

The aim of the experiments is to assess our pre-hoc and post-hoc approaches in terms of their performance for diagnosing malignancy and localising malignant lesions from breast DCE-MRI. 
Firstly, we individually evaluate the components of our proposed pre-hoc and post-hoc methods. 
Secondly, we compare the performance of both approaches in terms of diagnosis accuracy and malignant lesion localisation. 
Note that in every localisation evaluation we consider a region to be true positive if the Dice coefficient measured  between a candidate region and the ground truth lesion is at least 0.2~\citep{maicas2017deep,maicas2018lesion}. 

\subsubsection{Pre-hoc System}

The lesion detection step in the pre-hoc approach is evaluated in terms of the free response operating characteristic (FROC) curve measured patient-wisely (as in previous detection works~\citep{gubern2015automated,maicas2017deep,maicas2018lesion}), which compares the true positive rate (TPR) against the number of false positive detections per patient (FPP). We also measure the inference time in a computer with the following configuration: Intel Core i7, 12 GB of RAM and a GPU Nvidia Titan X 12 GB.
As in previous diagnosis work~\citep{maicas2017deep}, the diagnosis step in the pre-hoc method is evaluated in terms of the area under the receiving operation characteristic curve (AUC), which compares true positive diagnosis rate against false positive diagnosis rate. The AUC is measured breast-wise in two different scenarios: 1) all the breasts in the testing set are considered, and 2) only breasts with at least one detected region are considered. % 

The \textbf{lesion detection} uses a 3D ResNet trained from scratch with random bounding volumes sampled from the training volumes. More specifically, we sample 8000 positive and 8000 negative patches that are resized to $100\times100\times50$ (the input size to the 3D ResNet). The choice of the input size of the ResNet is $100\times100\times50$ so that every lesion is visible -- some tiny lesions disappear at finer resolutions. 
The architecture of the 3D ResNet~\citep{he2016deep} comprises 5 Residual Blocks~\citep{huang2016deep}, each of them preceded by a convolutional layer. 
After the last residual block, the model contains two additional convolutional layers and a fully connected (FC) layer.
The embedding of the observation ``$\mathbf{o}$'' is the output of the second to last convolutional layer, before the FC layer and it has 2304 dimensions.

The DQN is a 2-layer multi-layer perceptron, with each layer containing 512 nodes. It outputs the $Q$-value for 9 actions: translation by one third of the size of the observation in the positive or negative direction on each of the dimensions (i.e. 6 actions), scaling by one sixth of the size of the observation and is applied in every dimension (i.e. 2 actions) and the trigger action. The reward value for the trigger action has been empirically defined as $\eta=10$ if $\tau_w=0.2$ (i.e., the Dice coefficient is at least 0.2 during the trigger action), and the discount factor is $\gamma = 0.9$.
The DQN is trained with batches of 100 experiences from the experience replay memory $\mathcal{E}$, which can store $10000$ experiences. We use Adam optimizer~\citep{kingma2014adam} with a learning rate of $10^{-6}$.

During training, the model is initialized with one centred large observation covering 75\% of the input breast volume.
During inference, the lesion detection algorithm is launched from 13 different initializations in order to increase the chances of finding all possible lesions present in a breast.
In addition to the same initialization used during training, eight initializations are placed in each of the eight $50\times50\times25$ corner volumes, and four $50\times50\times25$ initializations are placed centred between the previous 8 initializations. 
The balance between exploration and exploitation during training is given by $\epsilon$, which decreases linearly from $\epsilon = 1$ to $\epsilon = 0.1$ after 300 epochs, and by $\kappa = 0.5$.

Detected regions are resized to $24\times24\times12$, which is the median value of the size of all detections in the training set. 
The \textbf{lesion diagnosis} uses a 3D DenseNet~\citep{huang2017densely} composed of three dense blocks of two dense layers each.
Each dense layer comprises a batch normalization, ReLu and a convolutional layer.
In the particular DenseNet implementation used in this paper, we use a compression of 0.5 and a growth rate of 6.
Global average pooling of $6\times6\times3$ is applied after the last dense block and before the fully connected layer.
The DenseNet is optimized with stochastic gradient descent with a learning rate of 0.01.
The dataset used to train the 3D DenseNet is composed of all detections obtained from the training set. Model selection is performed using the detections from the validation set based on the breast-wise AUC for breast screening.
Note that detections that correspond to malignant lesions are labelled as positive while detections that correspond to benign lesions or false positives are labelled as negative. 

\subsubsection{Post-hoc System}
\label{sec:densetOptimised}

The \textbf{diagnosis} step in the post-hoc approach is evaluated with the breast-wise AUC. 
The malignant lesion localization step in the post-hoc approach is evaluated in terms of FROC curve patient-wise % 
under two different scenarios: 1) all the patients in the test set are considered to compute the FROC (A), and 2) only the number of patients that had at least one breast diagnosed as malignant (+), such that the performance of the 1-class saliency detector can be isolated.

The breast volume  diagnosis meta-training algorithm uses as the underlying model a 3D DenseNet~\citep{huang2017densely}.
The architecture was decided based on the optimization of a 3D DenseNet (trained with the training set $\mathcal{T}$) to achieve the best results for the breast screening task on the validation set $\mathcal{V}$ and consists of 5 dense blocks with 2 dense layers each. Each dense layer comprises a batch normalization, ReLu and convolutional layer, where compression was 0.5 and growth rate 6. No data augmentation or dropout were used since they did not improve the performance of this 3D model.
For meta-training, the learning rate is $\alpha = 0.01$ and the meta-learning rate is $\beta = 0.001$. The number of gradient descent steps during adaptation is 5 and the number of meta-iterations is $M=3000$.
The meta-batch size contained $|\mathcal{K}_{t}| = 5$ tasks, 
where each task had $N^{tr} = 4$ samples for training and $ N^{val} = 4$ for validation. Each buffer $\mathcal{B}_j$  stored 40 recent rewards.

The localisation of \textbf{malignant lesions} in positively classified volumes is achieved by thresholding the generated saliency map at $\zeta = 0.8$ -- this threshold was decided based on the detection performance in the validation set.
The parameters for training the 1-class saliency detector in~\eqref{eq:PostweakLocLoss} are: $\lambda_1 = 0.1$, $\lambda_2 = 3$, $\lambda_3 = 1 $, and $\lambda_4 = 2.5$.

\subsubsection{Comparison Between Pre- and Post-Hoc}

Using the set-up described above for pre-hoc and post-hoc approaches, we compare the performance of both methods.
We evaluate diagnosis breast-wisely and patient-wisely in terms of the area under the receiving operation characteristic curve (AUC).
We also evaluate the performance of malignant lesion localisation for each approach patient-wisely using the FROC curve as in previous works~\citep{gubern2015automated,maicas2017deep,maicas2018lesion}. 
Note that the TPR for malignant lesion localization breast-wisely is the same as patient-wisely, while the FPR breast-wisely is the same as the one for patient-wisely divided by two.
For the post-hoc method, we also plot the two scenarios (A) and (+), explained above.

%%%%%%%%%%%%%%%%%%%%%%%%%%%%%%%%%%%%%%%%%%%%%	Experimental Reults
\subsubsection{Experimental Results for the Pre-hoc System}

\begin{figure}[h]
\centering\includegraphics[width=0.6\linewidth]{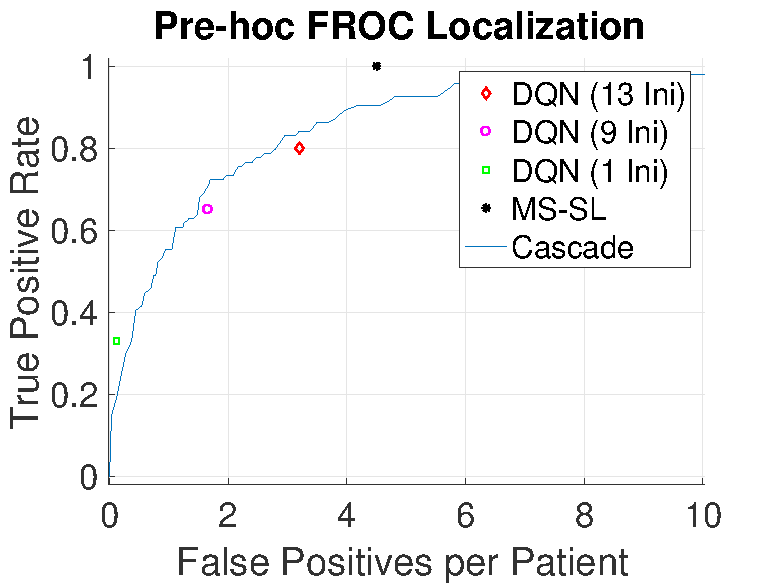}
\caption{FROC curve per patient for the lesion detection step of our pre-hoc method, labelled as DQN, where the information in brackets refers to the number of initialisations per breast used during the inference process.  MS-SL refers to the mean-shift structured learning approach, and Cascade denotes the multi-scale cascade method based on deep learning features.}
\label{fig:PreDet}
\end{figure}

We compare the performance of our \textbf{lesion detection} step against an improved version of exhaustive search, namely a multi-scale cascade based on deep learning features~\citep{maicas2017globally}, and a mean-shift clustering method followed by structured learning~\citep{mcclymont2014fully} (note that only one operating point is available for this approach), which is evaluated on the same dataset using a different training and testing data split.
Figure~\ref{fig:PreDet} shows the FROC curve with the detection results and Table~\ref{tab:preDetTimes} contains the inference times per patient needed by each of the methods.

\begin{table}[]
\begin{tabular}{l|c|}
\cline{2-2}
 & \multicolumn{1}{l|}{\textbf{Inference Time Per Patient}} \\ \hline
\multicolumn{1}{|l|}{\textbf{DQN ( 13 Initializations )}} & $92 \pm 21 s$ \\ \hline
\multicolumn{1}{|l|}{\textbf{MS-SL}} & $164 \pm 137 s$ \\ \hline
\multicolumn{1}{|l|}{\textbf{Cascade}} & $\mathcal{O}(60) min$ \\ \hline
\end{tabular}
\caption{Inference time per patient of our proposed pre-hoc detection method (DQN using 13 initializations per breast), the MS-SL (mean-shift structured learning), and the multi-scale cascade baselines.}
\label{tab:preDetTimes}
\end{table}

The \textbf{diagnosis} of breast volumes, based on the classification of the detected regions, achieves an AUC of 0.85, if all volumes in the test dataset are considered. 
%If we consider only the breast volumes, where there is at least one detected region, then the AUC increases to 0.86.

\subsubsection{Experimental Results for the Post-hoc System}

We evaluate the performance of our post-hoc diagnosis against three state-of-the-art classifiers.
The first baseline is the 3D DenseNet~\citep{huang2017densely} that has been optimized to solve the breast screening problem (as explained in Sec.~\ref{sec:densetOptimised}).
The second baseline is the same 3D DenseNet fine-tuned using a multiple instance learning (MIL) set-up~\citep{zhu2017deep}, which holds the state-of-the-art for the breast screening problem from mammography.
Finally, we compare against a 3D DenseNet trained from scratch using multi-task learning~\citep{xue2018full}, such that the model is jointly trained to solve all the tasks defined in Sec.~\ref{sec:VolDiagnose}. % 
See Table~\ref{tab:PostDiagnosis} for the AUC diagnosis results.

\begin{table}
\centering
\begin{tabular}{|l|c|}
\hline
\textbf{Baseline} & \textbf{AUC}                                  \\
\hline
Meta-Training\textbf{(Ours)}         & \textbf{0.90 }                   \\
\hline
Multi-task~\citep{xue2018full}         & 0.85                    \\
\hline
MIL~\citep{zhu2017deep}               & 0.85                                      \\
\hline
DenseNet~\citep{huang2017densely}          & 0.83                                           \\
\hline                
\end{tabular}
\caption{Breast-wise AUC for diagnosis in post-hoc systems. Our proposed post-hoc diagnosis method based on meta-training is labelled as Meta-Training, while the baseline based on multiple instance learning is labelled as MIL and the one based on multi-task learning is denoted as Multi-task.}
\label{tab:PostDiagnosis}
\end{table}

\begin{figure}[h]
\centering\includegraphics[width=0.6\linewidth]{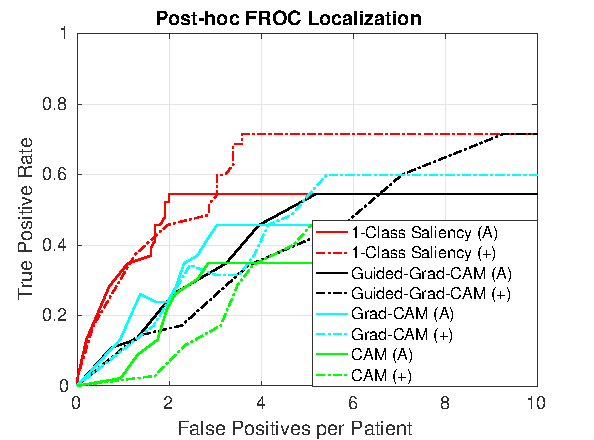}
\caption{Patient-wise FROC curves for post-hoc malignant lesion detection, where our method is denoted as 1-Class Saliency. Baselines are denoted as CAM~\citep{zhou2016learning}, and Grad-CAM and Guided Grad-CAM~\citep{selvaraju2017grad}. For each method, we present two scenarios: (A) all the volumes in the test set are considered to compute the FROC, and (+) only positively classified volumes are considered.}
\label{fig:PostDet}
\end{figure}

Our 1-class saliency detector specially designed to \textbf{detect malignant lesions} in positively classified volumes is compared against the following baselines: CAM~\citep{zhou2016learning}, and Grad-CAM and Guided Grad-CAM~\citep{selvaraju2017grad}. 
Figure~\ref{fig:PostDet} shows the FROC curves for our proposed methods and baselines in each of the two scenarios (A) and (+).

\subsubsection{Experimental Results for the Comparison Between Pre- and Post-Hoc}

\begin{figure}[h]
\centering\includegraphics[width=0.6\linewidth]{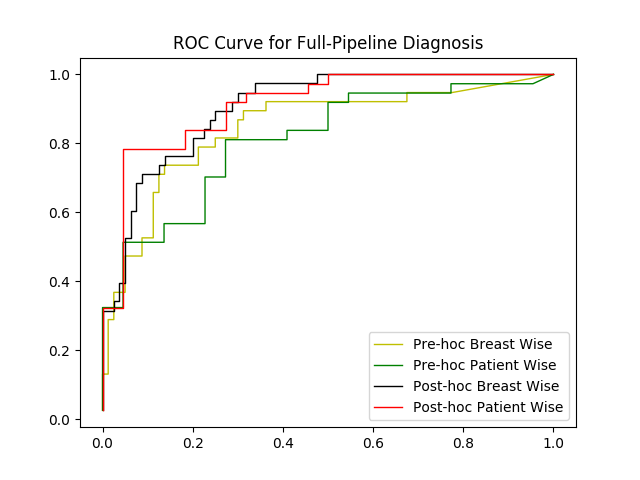}
\caption{ROC curves for malignancy diagnosis of pre-hoc and post-hoc full pipelines measures breast and patient-wise. }
\label{fig:fullDiag}
\end{figure}

\begin{figure}[h]
\centering\includegraphics[width=0.6\linewidth]{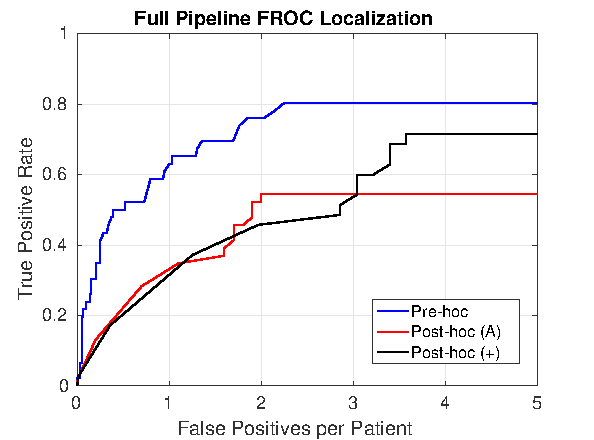}
\caption{Patient-wise FROC curve for malignant lesion detection of pre-hoc and post-hoc full pipeline methods. For the post-hoc method, we present two scenarios the two scenarios (A) and (+).}
\label{fig:fullDet}
\end{figure}

\begin{figure}[h]
\centering
\subfloat[]{\includegraphics[width=0.3\linewidth]{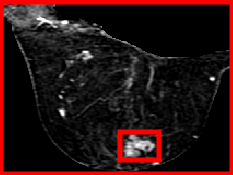}\label{img11}}\hspace{0.07in}
\subfloat[]{\includegraphics[width=0.3\linewidth]{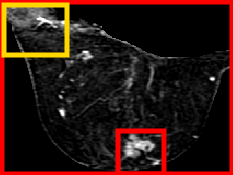}\label{img11}}\hspace{0.07in}
\subfloat[]{\includegraphics[width=0.3\linewidth]{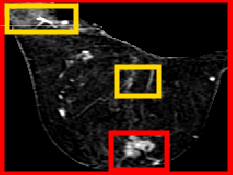}\label{img12}}\\
\subfloat[]{\includegraphics[width=0.3\linewidth]{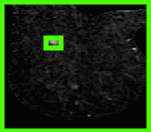}\label{img21}}\hspace{0.07in}
\subfloat[]{\includegraphics[width=0.3\linewidth]{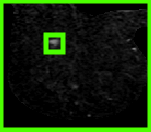}\label{img21}}\hspace{0.07in}
\subfloat[]{\includegraphics[width=0.3\linewidth]{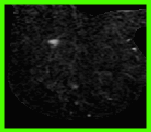}\label{img22}}\\
\caption{Example of two correct diagnosis by both pre-hoc and post-hoc full pipeline methods. Left column is the ground truth, middle column is the result of the pre-hoc method and right column is the result of the post-hoc method. Red image frames indicate malignant diagnosis, green frames indicate non-malignant diagnosis. Detections in red indicates TP malignant detections, yellow detections indicate FP malignant detections, detections in green indicate benign lesions.
\textbf{First row:} pre-hoc and post-hoc correct positive diagnosis with the malignant lesion detected.
\textbf{Second row:} pre-hoc and post-hoc correct negative diagnosis where the pre-hoc method correctly classified as negative a detected benign lesion and the post-hoc method did not localize any malignant lesion}
\label{fig:examplesFullCorrect}
\end{figure}

\begin{figure}[h]
\centering
\subfloat[]{\includegraphics[width=0.3\linewidth]{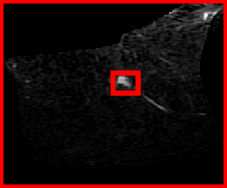}\label{img31}}\hspace{0.07in}
\subfloat[]{\includegraphics[width=0.3\linewidth]{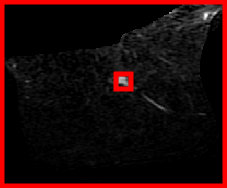}\label{img31}}\hspace{0.07in}
\subfloat[]{\includegraphics[width=0.3\linewidth]{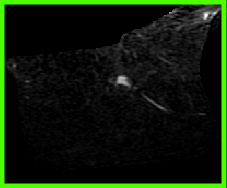}\label{img32}}\\
\subfloat[]{\includegraphics[width=0.3\linewidth]{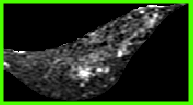}\label{img41}}\hspace{0.07in}
\subfloat[]{\includegraphics[width=0.3\linewidth]{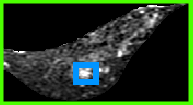}\label{img41}}\hspace{0.07in}
\subfloat[]{\includegraphics[width=0.3\linewidth]{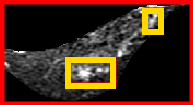}\label{img42}}\\
\caption{Example of two correct diagnosis by the pre-hoc system, but wrongly diagnosed by the post-hoc method. Left column is the ground truth, middle column is the result of the pre-hoc method and right column is the result of the post-hoc method. Red image frames indicate malignant diagnosis, green frames indicate non-malignant diagnosis. Detections in red indicate TP malignant detections, yellow detections indicate FP malignant detections, detection in blue indicates a ROI detection correctly classified as negative (non-malignant).
\textbf{First row:} correct positive diagnosis by the pre-hoc method with the malignant lesion correctly detected but incorrect non-malignant diagnosis by the post-hoc method.
\textbf{Second row:} correct negative diagnosis by the pre-hoc method, but incorrect positive diagnosis by the post-hoc system -- yielding the potential malignant regions in the rectangles shown in yellow. }% 
\label{fig:examplesFullPre}
\end{figure}

\begin{figure}[h]
\centering
\subfloat[]{\includegraphics[width=0.3\linewidth]{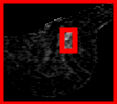}\label{img51}}\hspace{0.07in}
\subfloat[]{\includegraphics[width=0.3\linewidth]{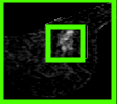}\label{img51}}\hspace{0.07in}
\subfloat[]{\includegraphics[width=0.3\linewidth]{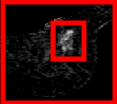}\label{img52}}\\
\subfloat[]{\includegraphics[width=0.3\linewidth]{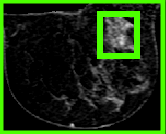}\label{img61}}\hspace{0.07in}
\subfloat[]{\includegraphics[width=0.3\linewidth]{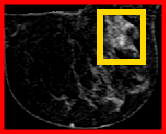}\label{img61}}\hspace{0.07in}
\subfloat[]{\includegraphics[width=0.3\linewidth]{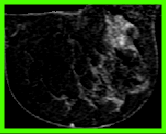}\label{img62}}\\
\caption{Example of two incorrect diagnosis by the pre-hoc, but correctly diagnosed by the post-hoc method. Left column is the ground truth, middle column is the result of the pre-hoc method and right column is the result of the post-hoc method. Red image frames indicate malignant diagnosis, green frames indicate non-malignant diagnosis. Detections in red indicate TP malignant detections, yellow detections indicate FP malignant detections, detection in green indicates a benign ROI detection.
\textbf{First Row:} the pre-hoc system incorrectly diagnoses as negative, while post-hoc system correctly diagnoses as positive and yields the malignant lesion.
\textbf{Second row:} the post-hoc method correctly diagnoses as negative, but pre-hoc incorrectly diagnoses as positive due to the wrong positive classification of a detected lesion} %w 
\label{fig:examplesFullPost}
\end{figure}

Table~\ref{tab:fullDiagnosis} contains the AUC for the malignancy diagnosis measured breast-wise and patient-wise for the pre-hoc and post-hoc approaches. Figure~\ref{fig:fullDiag} shows the ROC curves used in the computation of the AUC in Table~\ref{tab:fullDiagnosis}.
Figure~\ref{fig:fullDet} shows the FROC curves for malignant lesion detection of pre-hoc and  post-hoc ( (A) and (+) ) methods.
Figures~\ref{fig:examplesFullCorrect},~\ref{fig:examplesFullPre},  and~\ref{fig:examplesFullPost} display examples of breast diagnosis and lesion localizations obtained from the proposed pre-hoc and post-hoc methods, where both methods correctly performed diagnosis (Fig.~\ref{fig:examplesFullCorrect}), only the pre-hoc method correctly diagnosed the breast (Fig.~\ref{fig:examplesFullPre}), and only the post-hoc method correctly diagnosed the breast (Fig.~\ref{fig:examplesFullPost}).

\begin{table}[]
\centering
\begin{tabular}{c|c|c|}
\cline{2-3}
\multicolumn{1}{l|}{}              & Pre-Hoc & Post-Hoc \\ \hline
\multicolumn{1}{|c|}{Breast-wise}  &   0.85  &     \textbf{0.90} \\ \hline
\multicolumn{1}{|c|}{Patient-wise} &   0.81  &     \textbf{0.91} \\ \hline
\end{tabular}
\caption{AUC comparing the diagnosis performance between pre-hoc and post-hoc measured breast-wise and patient-wise.}
\label{tab:fullDiagnosis}
\end{table}

%%%%%%%%%%%%%%%%%%%%%%%%%%%%%%%%%%%%%%%%%%%%%	Discussion
\section{Discussion}
\label{S:5}

The localization step in the pre-hoc method achieves similar accuracy to the baseline methods. As shown in Figure~\ref{fig:PreDet}, the TPR and FPR directly depends on the number of initializations used by the reinforcement learning algorithm. %I 
In addition, the performance of our localization step is very similar to the baseline based on a multi-scale cascade using exhaustive search with deep features. However, multi-scale cascade (164s) and clustering+structure learning (several hours) methods require large inference times compared to our attention model (92s) as shown in Table~\ref{tab:preDetTimes}.

The post-hoc diagnosis step improves over several baseline methods, as shown in Table~\ref{tab:PostDiagnosis}. 
These baseline methods are based on a DenseNet~\citep{huang2017densely}, specifically optimised for the breast screening classification, and on extensions derived from multiple instance learning~\citep{zhu2017deep} and multi-task learning~\citep{xue2018full}. 
These results show that meta-training the model to solve tasks with small training sets is an important step to improve the learning of methods when only small datasets are available. Baseline approaches~\citep{huang2017densely,xue2018full} only show a limited improvement over the DenseNet baseline.

The localization step in our post-hoc method benefits from our definition of saliency, as shown in Figure.~\ref{fig:PostDet}.  In contrast, baseline methods show activations that do not correlate well with the target classification. In addition, baseline methods, such as CAM~\citep{zhou2016learning} and Grad-Cam~\citep{selvaraju2017grad}, suffer from the low resolution of the activation feature maps, despite the improvement achieved by Guided Grad-Cam~\citep{selvaraju2017grad}.
Measuring results only on positively classified volumes ( (+) curves in Figure~\ref{fig:PostDet}) discounts the mistakes made by the diagnosis step and provides an evaluation that isolates the lesion detection ( (A) curves in Figure~\ref{fig:PostDet}).
Note that there is no straightforward comparison between the localization steps in post-hoc methodologies (that only detects malignant lesions) and the localization step in pre-hoc methodologies (that detects benign and malignant lesions). Such malignant lesion detection comparison only makes sense in terms of the full pre-hoc and post-hoc pipelines, which is detailed below.

In terms of the full pipeline, we observe in Table~\ref{tab:fullDiagnosis} and Figure~\ref{fig:fullDiag} that the post-hoc system has a higher classification AUC than the pre-hoc for breast screening from breast DCE-MRI. The difference between these two methods is higher when measured patient-wise compared to breast-wise. It seems reasonable to think that the reason behind such discrepancy is an effect of the missed detections in pre-hoc. In difficult (small and low contrast lesions) cases with missed detections, the confidence score of malignancy of a breast is considered 0. While this effect is smaller when the AUC is measured breast-wise (as there are 118 samples of breasts), it is larger when measured patient-wise (59 samples of patients). Furthermore, the better results of the post-hoc method suggest that the analysis of the whole image allows it to find indications for malignancy that are located in other areas of the image~\citep{kostopoulos2017computer}.

Regarding the localization of malignant lesions, the pre-hoc system achieves better accuracy, compared with the post-hoc. 
This suggests that the strong annotations used to train the pre-hoc method gives it an advantage for the localisation of lesions, when compared with the weak annotation used to train the post-hoc approach. This issue is exemplified in Figure~\ref{fig:examplesFullCorrect} (Row 1), where although both approaches present a correct diagnosis, the post-hoc method yields a higher number of false positive malignant lesion detections. A similar behaviour can be seen in Figure~\ref{fig:examplesFullPre} (Row 2), where the post-hoc produces an incorrect diagnosis and additionally yields two false positive detections. 
In addition, the detection step for the pre-hoc system is mainly designed to achieve good performance when only a small training set is available. On the contrary, the malignant lesion localization step in the post-hoc approach is not particularly focused on being able to perform well from a small dataset. This difference in design focus is likely to be influencing the detection results too. % 
Finally, it is worth noting that the FROC for the post-hoc approach (Post-hoc (A) curve in Fig.~\ref{fig:fullDet}) is affected by the diagnosis process. However, if we remove the effect of the diagnosis step and consider the performance of malignant lesion localization in positively classified volumes, we observe a closer performance compared to the pre-hoc method, even though the post-hoc system is trained with weak annotations.

%%%%%%%%%%%%%%%%%%%%%%%%%%%%%%%%%%%%%%%%%%%%%	Future Work
\section{Limitations and Future Work}
\label{S:6}

The main limitation of our work comes from the small dataset available. In addition to a larger test set, we aim to increase our dataset to include patients where no lesions are found in order to better recreate the scenario of a screening population. Ideally, this dataset will contain scanners from different vendors too. 
Another limitation of this work involves the lack of cross-validation experiments. This decision is justified to allow a fair comparison with other works~\citep{maicas2017globally,maicas2017deep,maicas2018training,maicas2018lesion,mcclymont2014fully} on the same dataset.

Future work involves the improvement of the malignant lesion localization in post-hoc methodologies by designing a new method specifically for the small training set available. 
We believe that the Lesion localization step in pre-hoc approaches could also be improved in terms of inference time and accuracy. As noted in~\citep{maicas2017deep}, improvements in running time can be achieved by running the different initializations of the detection algorithm in parallel and by optimizing the resizing operation of the current bounding volume.
In addition, the use of a U-net~\citep{ronneberger2015u} would allow the implementation of a faster segmentation map maintaining the detection accuracy.
Finally, it would be interesting to design a method that could diagnose based on the combined analysis of MRI and mammography.

%%%%%%%%%%%%%%%%%%%%%%%%%%%%%%%%%%%%%%%%%%%%%	Conclusion
\section{Conclusion}
\label{S:7}

We introduced and compared two different approaches for breast screening from breast DCE-MRI: pre-hoc and post-hoc methods. 
The pre-hoc method localizes suspicious regions (benign and malignant lesions) using an attention model based on deep reinforcement learning.
Detected regions were subsequently classified into malignant or non-malignant lesions using a 3D DenseNet. 
The post-hoc method diagnoses a DCE-MRI breast volume using a classifier that, before being trained to solve the breast screening task, has been  meta-trained to solve several breast-related tasks where only small training sets are available. 
Malignant regions are then localized with a 1-class saliency detector specifically designed for post-hoc systems that perform diagnosis.
Results showed that the post-hoc method can achieve better performance for malignancy diagnosis, whereas the pre-hoc method could more precisely localize malignant lesions. However, this improvement of the pre-hoc detection method relies on the employment of  strong annotations during the training process. 
On the other hand, post-hoc  methods only use weak labels during the training phase and outperforms pre-hoc methods in diagnosis, which is the main aim of a breast screening system.
In conclusion, we believe that future research should focus on the development and improvement of post-hoc diagnosis methods.

We would like to thank Nvidia for the donation of a TitanXp that supported this work. 
%%%%%%%%%%%%%%%%%%%%%%%%%%%%%%%%%%%%%%%%%%%%%	END



%% References


%\bibliographystyle{model4-names}
%\bibliographystyle{elsarticle-num}
\bibliographystyle{model2-names.bst}\biboptions{authoryear}%OK
%\bibliographystyle{model1-num-names.bst}%\biboptions{authoryear}
%%GABI: \bibliographystyle{model1-num-names} %From previous
\section*{References}%Gabi includes
\bibliography{sample.bib}

\begin{thebibliography}{73}
\expandafter\ifx\csname natexlab\endcsname\relax\def\natexlab#1{#1}\fi
\providecommand{\url}[1]{\texttt{#1}}
\providecommand{\href}[2]{#2}
\providecommand{\path}[1]{#1}
\providecommand{\DOIprefix}{doi:}
\providecommand{\ArXivprefix}{arXiv:}
\providecommand{\URLprefix}{URL: }
\providecommand{\Pubmedprefix}{pmid:}
\providecommand{\doi}[1]{\href{http://dx.doi.org/#1}{\path{#1}}}
\providecommand{\Pubmed}[1]{\href{pmid:#1}{\path{#1}}}
\providecommand{\bibinfo}[2]{#2}
\ifx\xfnm\relax \def\xfnm[#1]{\unskip,\space#1}\fi
%Type = Article
\bibitem[{Agner et~al.(2014)Agner, Rosen, Englander, Tomaszewski, Feldman,
  Zhang, Mies, Schnall and Madabhushi}]{agner2014computerized}
\bibinfo{author}{Agner, S.C.}, \bibinfo{author}{Rosen, M.A.},
  \bibinfo{author}{Englander, S.}, \bibinfo{author}{Tomaszewski, J.E.},
  \bibinfo{author}{Feldman, M.D.}, \bibinfo{author}{Zhang, P.},
  \bibinfo{author}{Mies, C.}, \bibinfo{author}{Schnall, M.D.},
  \bibinfo{author}{Madabhushi, A.}, \bibinfo{year}{2014}.
\newblock \bibinfo{title}{Computerized image analysis for identifying
  triple-negative breast cancers and differentiating them from other molecular
  subtypes of breast cancer on dynamic contrast-enhanced mr images: a
  feasibility study}.
\newblock \bibinfo{journal}{Radiology} \bibinfo{volume}{272},
  \bibinfo{pages}{91--99}.
%Type = Techreport
\bibitem[{AIHW(2007)}]{ausStats2017}
\bibinfo{author}{AIHW}, \bibinfo{year}{2007}.
\newblock \bibinfo{title}{Cancer in Australia 2017}.
\newblock \bibinfo{type}{Technical Report}. The Australian Institute of Health
  and Welfare.
%Type = Inproceedings
\bibitem[{Amit et~al.(2017a)Amit, Ben-Ari, Hadad, Monovich, Granot and
  Hashoul}]{amit2017classification}
\bibinfo{author}{Amit, G.}, \bibinfo{author}{Ben-Ari, R.},
  \bibinfo{author}{Hadad, O.}, \bibinfo{author}{Monovich, E.},
  \bibinfo{author}{Granot, N.}, \bibinfo{author}{Hashoul, S.},
  \bibinfo{year}{2017}a.
\newblock \bibinfo{title}{Classification of breast mri lesions using small-size
  training sets: comparison of deep learning approaches}, in:
  \bibinfo{booktitle}{Medical Imaging 2017: Computer-Aided Diagnosis},
  \bibinfo{organization}{International Society for Optics and Photonics}. p.
  \bibinfo{pages}{101341H}.
%Type = Inproceedings
\bibitem[{Amit et~al.(2017b)Amit, Hadad, Alpert, Tlusty, Gur, Ben-Ari and
  Hashoul}]{amit2017hybrid}
\bibinfo{author}{Amit, G.}, \bibinfo{author}{Hadad, O.},
  \bibinfo{author}{Alpert, S.}, \bibinfo{author}{Tlusty, T.},
  \bibinfo{author}{Gur, Y.}, \bibinfo{author}{Ben-Ari, R.},
  \bibinfo{author}{Hashoul, S.}, \bibinfo{year}{2017}b.
\newblock \bibinfo{title}{Hybrid mass detection in breast mri combining
  unsupervised saliency analysis and deep learning}, in:
  \bibinfo{booktitle}{International Conference on Medical Image Computing and
  Computer-Assisted Intervention}, \bibinfo{organization}{Springer}. pp.
  \bibinfo{pages}{594--602}.
%Type = Article
\bibitem[{Behrens et~al.(2007)Behrens, Laue, Althaus, Boehler, Kuemmerlen, Hahn
  and Peitgen}]{behrens2007computer}
\bibinfo{author}{Behrens, S.}, \bibinfo{author}{Laue, H.},
  \bibinfo{author}{Althaus, M.}, \bibinfo{author}{Boehler, T.},
  \bibinfo{author}{Kuemmerlen, B.}, \bibinfo{author}{Hahn, H.K.},
  \bibinfo{author}{Peitgen, H.O.}, \bibinfo{year}{2007}.
\newblock \bibinfo{title}{Computer assistance for mr based diagnosis of breast
  cancer: present and future challenges}.
\newblock \bibinfo{journal}{Computerized medical imaging and graphics}
  \bibinfo{volume}{31}, \bibinfo{pages}{236--247}.
%Type = Inproceedings
\bibitem[{Caicedo and Lazebnik(2015)}]{caicedo2015active}
\bibinfo{author}{Caicedo, J.C.}, \bibinfo{author}{Lazebnik, S.},
  \bibinfo{year}{2015}.
\newblock \bibinfo{title}{Active object localization with deep reinforcement
  learning}, in: \bibinfo{booktitle}{Proceedings of the IEEE International
  Conference on Computer Vision}, pp. \bibinfo{pages}{2488--2496}.
%Type = Inproceedings
\bibitem[{Caruana et~al.(2015)Caruana, Lou, Gehrke, Koch, Sturm and
  Elhadad}]{caruana2015intelligible}
\bibinfo{author}{Caruana, R.}, \bibinfo{author}{Lou, Y.},
  \bibinfo{author}{Gehrke, J.}, \bibinfo{author}{Koch, P.},
  \bibinfo{author}{Sturm, M.}, \bibinfo{author}{Elhadad, N.},
  \bibinfo{year}{2015}.
\newblock \bibinfo{title}{Intelligible models for healthcare: Predicting
  pneumonia risk and hospital 30-day readmission}, in:
  \bibinfo{booktitle}{Proceedings of the 21th ACM SIGKDD International
  Conference on Knowledge Discovery and Data Mining},
  \bibinfo{organization}{ACM}. pp. \bibinfo{pages}{1721--1730}.
%Type = Article
\bibitem[{Chen et~al.(2006)Chen, Giger and Bick}]{chen2006fuzzy}
\bibinfo{author}{Chen, W.}, \bibinfo{author}{Giger, M.L.},
  \bibinfo{author}{Bick, U.}, \bibinfo{year}{2006}.
\newblock \bibinfo{title}{A fuzzy c-means (fcm)-based approach for computerized
  segmentation of breast lesions in dynamic contrast-enhanced mr images}.
\newblock \bibinfo{journal}{Academic radiology} \bibinfo{volume}{13},
  \bibinfo{pages}{63--72}.
%Type = Article
\bibitem[{Dalm{\i}{\c{s}} et~al.(2016)Dalm{\i}{\c{s}}, Gubern-M{\'e}rida,
  Vreemann, Karssemeijer, Mann and Platel}]{dalmics2016computer}
\bibinfo{author}{Dalm{\i}{\c{s}}, M.U.}, \bibinfo{author}{Gubern-M{\'e}rida,
  A.}, \bibinfo{author}{Vreemann, S.}, \bibinfo{author}{Karssemeijer, N.},
  \bibinfo{author}{Mann, R.}, \bibinfo{author}{Platel, B.},
  \bibinfo{year}{2016}.
\newblock \bibinfo{title}{A computer-aided diagnosis system for breast dce-mri
  at high spatiotemporal resolution}.
\newblock \bibinfo{journal}{Medical physics} \bibinfo{volume}{43},
  \bibinfo{pages}{84--94}.
%Type = Article
\bibitem[{Dalm{\i}{\c{s}} et~al.(2018)Dalm{\i}{\c{s}}, Vreemann, Kooi, Mann,
  Karssemeijer and Gubern-M{\'e}rida}]{dalmics2018fully}
\bibinfo{author}{Dalm{\i}{\c{s}}, M.U.}, \bibinfo{author}{Vreemann, S.},
  \bibinfo{author}{Kooi, T.}, \bibinfo{author}{Mann, R.M.},
  \bibinfo{author}{Karssemeijer, N.}, \bibinfo{author}{Gubern-M{\'e}rida, A.},
  \bibinfo{year}{2018}.
\newblock \bibinfo{title}{Fully automated detection of breast cancer in
  screening mri using convolutional neural networks}.
\newblock \bibinfo{journal}{Journal of Medical Imaging} \bibinfo{volume}{5},
  \bibinfo{pages}{014502}.
%Type = Article
\bibitem[{DeSantis et~al.(2015)DeSantis, Bray, Ferlay, Lortet-Tieulent,
  Anderson and Jemal}]{desantis2015international}
\bibinfo{author}{DeSantis, C.E.}, \bibinfo{author}{Bray, F.},
  \bibinfo{author}{Ferlay, J.}, \bibinfo{author}{Lortet-Tieulent, J.},
  \bibinfo{author}{Anderson, B.O.}, \bibinfo{author}{Jemal, A.},
  \bibinfo{year}{2015}.
\newblock \bibinfo{title}{International variation in female breast cancer
  incidence and mortality rates}.
\newblock \bibinfo{journal}{Cancer Epidemiology and Prevention Biomarkers}
  \bibinfo{volume}{24}, \bibinfo{pages}{1495--1506}.
%Type = Inproceedings
\bibitem[{Dubost et~al.(2017)Dubost, Bortsova, Adams, Ikram, Niessen, Vernooij
  and De~Bruijne}]{dubost2017gp}
\bibinfo{author}{Dubost, F.}, \bibinfo{author}{Bortsova, G.},
  \bibinfo{author}{Adams, H.}, \bibinfo{author}{Ikram, A.},
  \bibinfo{author}{Niessen, W.J.}, \bibinfo{author}{Vernooij, M.},
  \bibinfo{author}{De~Bruijne, M.}, \bibinfo{year}{2017}.
\newblock \bibinfo{title}{Gp-unet: Lesion detection from weak labels with a 3d
  regression network}, in: \bibinfo{booktitle}{International Conference on
  Medical Image Computing and Computer-Assisted Intervention},
  \bibinfo{organization}{Springer}. pp. \bibinfo{pages}{214--221}.
%Type = Article
\bibitem[{Esteva et~al.(2017)Esteva, Kuprel, Novoa, Ko, Swetter, Blau and
  Thrun}]{esteva2017dermatologist}
\bibinfo{author}{Esteva, A.}, \bibinfo{author}{Kuprel, B.},
  \bibinfo{author}{Novoa, R.A.}, \bibinfo{author}{Ko, J.},
  \bibinfo{author}{Swetter, S.M.}, \bibinfo{author}{Blau, H.M.},
  \bibinfo{author}{Thrun, S.}, \bibinfo{year}{2017}.
\newblock \bibinfo{title}{Dermatologist-level classification of skin cancer
  with deep neural networks}.
\newblock \bibinfo{journal}{Nature} .
%Type = Inproceedings
\bibitem[{Feng et~al.(2017)Feng, Yang, Laine and
  Angelini}]{feng2017discriminative}
\bibinfo{author}{Feng, X.}, \bibinfo{author}{Yang, J.}, \bibinfo{author}{Laine,
  A.F.}, \bibinfo{author}{Angelini, E.D.}, \bibinfo{year}{2017}.
\newblock \bibinfo{title}{Discriminative localization in cnns for
  weakly-supervised segmentation of pulmonary nodules}, in:
  \bibinfo{booktitle}{International Conference on Medical Image Computing and
  Computer-Assisted Intervention}, \bibinfo{organization}{Springer}. pp.
  \bibinfo{pages}{568--576}.
%Type = Article
\bibitem[{Gallego-Ortiz and Martel(2015)}]{gallego2015improving}
\bibinfo{author}{Gallego-Ortiz, C.}, \bibinfo{author}{Martel, A.L.},
  \bibinfo{year}{2015}.
\newblock \bibinfo{title}{Improving the accuracy of computer-aided diagnosis
  for breast mr imaging by differentiating between mass and nonmass lesions}.
\newblock \bibinfo{journal}{Radiology} \bibinfo{volume}{278},
  \bibinfo{pages}{679--688}.
%Type = Article
\bibitem[{Gilbert and Selamoglu(2018)}]{gilbert2018personalised}
\bibinfo{author}{Gilbert, F.}, \bibinfo{author}{Selamoglu, A.},
  \bibinfo{year}{2018}.
\newblock \bibinfo{title}{Personalised screening: is this the way forward?}
\newblock \bibinfo{journal}{Clinical radiology} \bibinfo{volume}{73},
  \bibinfo{pages}{327--333}.
%Type = Article
\bibitem[{Grimm et~al.(2015)Grimm, Anderson, Baker, Johnson, Walsh, Yoon and
  Ghate}]{grimm2015interobserver}
\bibinfo{author}{Grimm, L.J.}, \bibinfo{author}{Anderson, A.L.},
  \bibinfo{author}{Baker, J.A.}, \bibinfo{author}{Johnson, K.S.},
  \bibinfo{author}{Walsh, R.}, \bibinfo{author}{Yoon, S.C.},
  \bibinfo{author}{Ghate, S.V.}, \bibinfo{year}{2015}.
\newblock \bibinfo{title}{Interobserver variability between breast imagers
  using the fifth edition of the bi-rads mri lexicon}.
\newblock \bibinfo{journal}{American Journal of Roentgenology}
  \bibinfo{volume}{204}, \bibinfo{pages}{1120--1124}.
%Type = Article
\bibitem[{Gubern-M{\'e}rida et~al.(2015)Gubern-M{\'e}rida, Mart{\'\i},
  Melendez, Hauth, Mann, Karssemeijer and Platel}]{gubern2015automated}
\bibinfo{author}{Gubern-M{\'e}rida, A.}, \bibinfo{author}{Mart{\'\i}, R.},
  \bibinfo{author}{Melendez, J.}, \bibinfo{author}{Hauth, J.L.},
  \bibinfo{author}{Mann, R.M.}, \bibinfo{author}{Karssemeijer, N.},
  \bibinfo{author}{Platel, B.}, \bibinfo{year}{2015}.
\newblock \bibinfo{title}{Automated localization of breast cancer in dce-mri}.
\newblock \bibinfo{journal}{Medical image analysis} \bibinfo{volume}{20},
  \bibinfo{pages}{265--274}.
%Type = Article
\bibitem[{Gubern-M{\'e}rida et~al.(2016)Gubern-M{\'e}rida, Vreemann,
  Mart{\'\i}, Melendez, Lardenoije, Mann, Karssemeijer and
  Platel}]{gubern2016automated}
\bibinfo{author}{Gubern-M{\'e}rida, A.}, \bibinfo{author}{Vreemann, S.},
  \bibinfo{author}{Mart{\'\i}, R.}, \bibinfo{author}{Melendez, J.},
  \bibinfo{author}{Lardenoije, S.}, \bibinfo{author}{Mann, R.M.},
  \bibinfo{author}{Karssemeijer, N.}, \bibinfo{author}{Platel, B.},
  \bibinfo{year}{2016}.
\newblock \bibinfo{title}{Automated detection of breast cancer in
  false-negative screening mri studies from women at increased risk}.
\newblock \bibinfo{journal}{European journal of radiology}
  \bibinfo{volume}{85}, \bibinfo{pages}{472--479}.
%Type = Article
\bibitem[{Hayton et~al.(1997)Hayton, Brady, Tarassenko and
  Moore}]{hayton1997analysis}
\bibinfo{author}{Hayton, P.}, \bibinfo{author}{Brady, M.},
  \bibinfo{author}{Tarassenko, L.}, \bibinfo{author}{Moore, N.},
  \bibinfo{year}{1997}.
\newblock \bibinfo{title}{Analysis of dynamic mr breast images using a model of
  contrast enhancement}.
\newblock \bibinfo{journal}{Medical image analysis} \bibinfo{volume}{1},
  \bibinfo{pages}{207--224}.
%Type = Inproceedings
\bibitem[{He et~al.(2017)He, Gkioxari, Doll{\'a}r and Girshick}]{he2017mask}
\bibinfo{author}{He, K.}, \bibinfo{author}{Gkioxari, G.},
  \bibinfo{author}{Doll{\'a}r, P.}, \bibinfo{author}{Girshick, R.},
  \bibinfo{year}{2017}.
\newblock \bibinfo{title}{Mask r-cnn}, in: \bibinfo{booktitle}{Computer Vision
  (ICCV), 2017 IEEE International Conference on}, \bibinfo{organization}{IEEE}.
  pp. \bibinfo{pages}{2980--2988}.
%Type = Inproceedings
\bibitem[{He et~al.(2016)He, Zhang, Ren and Sun}]{he2016deep}
\bibinfo{author}{He, K.}, \bibinfo{author}{Zhang, X.}, \bibinfo{author}{Ren,
  S.}, \bibinfo{author}{Sun, J.}, \bibinfo{year}{2016}.
\newblock \bibinfo{title}{Deep residual learning for image recognition}, in:
  \bibinfo{booktitle}{Proceedings of the IEEE conference on computer vision and
  pattern recognition}, pp. \bibinfo{pages}{770--778}.
%Type = Inproceedings
\bibitem[{Huang et~al.(2017)Huang, Liu, van~der Maaten and
  Weinberger}]{huang2017densely}
\bibinfo{author}{Huang, G.}, \bibinfo{author}{Liu, Z.},
  \bibinfo{author}{van~der Maaten, L.}, \bibinfo{author}{Weinberger, K.Q.},
  \bibinfo{year}{2017}.
\newblock \bibinfo{title}{Densely connected convolutional networks}, in:
  \bibinfo{booktitle}{Proceedings of the IEEE Conference on Computer Vision and
  Pattern Recognition}, pp. \bibinfo{pages}{4700--4708}.
%Type = Inproceedings
\bibitem[{Huang et~al.(2016)Huang, Sun, Liu, Sedra and
  Weinberger}]{huang2016deep}
\bibinfo{author}{Huang, G.}, \bibinfo{author}{Sun, Y.}, \bibinfo{author}{Liu,
  Z.}, \bibinfo{author}{Sedra, D.}, \bibinfo{author}{Weinberger, K.Q.},
  \bibinfo{year}{2016}.
\newblock \bibinfo{title}{Deep networks with stochastic depth}, in:
  \bibinfo{booktitle}{European Conference on Computer Vision},
  \bibinfo{organization}{Springer}. pp. \bibinfo{pages}{646--661}.
%Type = Article
\bibitem[{Kingma and Ba(2014)}]{kingma2014adam}
\bibinfo{author}{Kingma, D.}, \bibinfo{author}{Ba, J.}, \bibinfo{year}{2014}.
\newblock \bibinfo{title}{Adam: A method for stochastic optimization}.
\newblock \bibinfo{journal}{arXiv preprint arXiv:1412.6980} .
%Type = Article
\bibitem[{Kostopoulos et~al.(2017)Kostopoulos, Vassiou, Lavdas, Cavouras,
  Kalatzis, Asvestas, Arvanitis, Fezoulidis and
  Glotsos}]{kostopoulos2017computer}
\bibinfo{author}{Kostopoulos, S.A.}, \bibinfo{author}{Vassiou, K.G.},
  \bibinfo{author}{Lavdas, E.N.}, \bibinfo{author}{Cavouras, D.A.},
  \bibinfo{author}{Kalatzis, I.K.}, \bibinfo{author}{Asvestas, P.A.},
  \bibinfo{author}{Arvanitis, D.L.}, \bibinfo{author}{Fezoulidis, I.V.},
  \bibinfo{author}{Glotsos, D.T.}, \bibinfo{year}{2017}.
\newblock \bibinfo{title}{Computer-based automated estimation of breast
  vascularity and correlation with breast cancer in dce-mri images}.
\newblock \bibinfo{journal}{Magnetic resonance imaging} \bibinfo{volume}{35},
  \bibinfo{pages}{39--45}.
%Type = Article
\bibitem[{Kousi et~al.(2015)Kousi, Borri, Dean, Panek, Scurr, Leach and
  Schmidt}]{kousi2015quality}
\bibinfo{author}{Kousi, E.}, \bibinfo{author}{Borri, M.},
  \bibinfo{author}{Dean, J.}, \bibinfo{author}{Panek, R.},
  \bibinfo{author}{Scurr, E.}, \bibinfo{author}{Leach, M.O.},
  \bibinfo{author}{Schmidt, M.A.}, \bibinfo{year}{2015}.
\newblock \bibinfo{title}{Quality assurance in mri breast screening: comparing
  signal-to-noise ratio in dynamic contrast-enhanced imaging protocols}.
\newblock \bibinfo{journal}{Physics in Medicine \& Biology}
  \bibinfo{volume}{61}, \bibinfo{pages}{37}.
%Type = Article
\bibitem[{Kriege et~al.(2004)Kriege, Brekelmans, Boetes, Besnard, Zonderland,
  Obdeijn, Manoliu, Kok, Peterse, Tilanus-Linthorst
  et~al.}]{kriege2004efficacy}
\bibinfo{author}{Kriege, M.}, \bibinfo{author}{Brekelmans, C.T.},
  \bibinfo{author}{Boetes, C.}, \bibinfo{author}{Besnard, P.E.},
  \bibinfo{author}{Zonderland, H.M.}, \bibinfo{author}{Obdeijn, I.M.},
  \bibinfo{author}{Manoliu, R.A.}, \bibinfo{author}{Kok, T.},
  \bibinfo{author}{Peterse, H.}, \bibinfo{author}{Tilanus-Linthorst, M.M.},
  et~al., \bibinfo{year}{2004}.
\newblock \bibinfo{title}{Efficacy of mri and mammography for breast-cancer
  screening in women with a familial or genetic predisposition}.
\newblock \bibinfo{journal}{New England Journal of Medicine}
  \bibinfo{volume}{351}, \bibinfo{pages}{427--437}.
%Type = Inproceedings
\bibitem[{Krizhevsky et~al.(2012)Krizhevsky, Sutskever and
  Hinton}]{krizhevsky2012imagenet}
\bibinfo{author}{Krizhevsky, A.}, \bibinfo{author}{Sutskever, I.},
  \bibinfo{author}{Hinton, G.E.}, \bibinfo{year}{2012}.
\newblock \bibinfo{title}{Imagenet classification with deep convolutional
  neural networks}, in: \bibinfo{booktitle}{Advances in neural information
  processing systems}, pp. \bibinfo{pages}{1097--1105}.
%Type = Article
\bibitem[{Lehman et~al.(2013)Lehman, Blume, DeMartini, Hylton, Herman and
  Schnall}]{lehman2013accuracy}
\bibinfo{author}{Lehman, C.D.}, \bibinfo{author}{Blume, J.D.},
  \bibinfo{author}{DeMartini, W.B.}, \bibinfo{author}{Hylton, N.M.},
  \bibinfo{author}{Herman, B.}, \bibinfo{author}{Schnall, M.D.},
  \bibinfo{year}{2013}.
\newblock \bibinfo{title}{Accuracy and interpretation time of computer-aided
  detection among novice and experienced breast mri readers}.
\newblock \bibinfo{journal}{American Journal of Roentgenology}
  \bibinfo{volume}{200}, \bibinfo{pages}{W683--W689}.
%Type = Article
\bibitem[{Levman et~al.(2009)Levman, Causer, Warner and
  Martel}]{levman2009effect}
\bibinfo{author}{Levman, J.E.}, \bibinfo{author}{Causer, P.},
  \bibinfo{author}{Warner, E.}, \bibinfo{author}{Martel, A.L.},
  \bibinfo{year}{2009}.
\newblock \bibinfo{title}{Effect of the enhancement threshold on the
  computer-aided detection of breast cancer using mri}.
\newblock \bibinfo{journal}{Academic radiology} \bibinfo{volume}{16},
  \bibinfo{pages}{1064--1069}.
%Type = Inproceedings
\bibitem[{Li et~al.(2018)Li, Wang, Han, Xue, Wei, Li and
  Fei-Fei}]{li2018thoracic}
\bibinfo{author}{Li, Z.}, \bibinfo{author}{Wang, C.}, \bibinfo{author}{Han,
  M.}, \bibinfo{author}{Xue, Y.}, \bibinfo{author}{Wei, W.},
  \bibinfo{author}{Li, L.J.}, \bibinfo{author}{Fei-Fei, L.},
  \bibinfo{year}{2018}.
\newblock \bibinfo{title}{Thoracic disease identification and localization with
  limited supervision}, in: \bibinfo{booktitle}{Proceedings of the IEEE
  conference on computer vision and pattern recognition}.
%Type = Article
\bibitem[{Liu et~al.(2017)Liu, Zheng, Liang, Tang, Ren, Zhang and
  Zhao}]{liu2017total}
\bibinfo{author}{Liu, H.}, \bibinfo{author}{Zheng, Y.}, \bibinfo{author}{Liang,
  D.}, \bibinfo{author}{Tang, P.}, \bibinfo{author}{Ren, F.},
  \bibinfo{author}{Zhang, L.}, \bibinfo{author}{Zhao, Z.},
  \bibinfo{year}{2017}.
\newblock \bibinfo{title}{Total variation based dce-mri decomposition by
  separating lesion from background for time-intensity curve estimation}.
\newblock \bibinfo{journal}{Medical physics} \bibinfo{volume}{44},
  \bibinfo{pages}{2321--2331}.
%Type = Inproceedings
\bibitem[{Maicas et~al.(2018a)Maicas, Bradley, Nascimento, Reid and
  Carneiro}]{maicas2018training}
\bibinfo{author}{Maicas, G.}, \bibinfo{author}{Bradley, A.P.},
  \bibinfo{author}{Nascimento, J.C.}, \bibinfo{author}{Reid, I.},
  \bibinfo{author}{Carneiro, G.}, \bibinfo{year}{2018}a.
\newblock \bibinfo{title}{Training medical image analysis systems like
  radiologists}, in: \bibinfo{booktitle}{Medical Image Computing and Computer
  Assisted Intervention -- MICCAI 2018}, \bibinfo{publisher}{Springer
  International Publishing}, \bibinfo{address}{Cham}. pp.
  \bibinfo{pages}{546--554}.
%Type = Inproceedings
\bibitem[{Maicas et~al.(2017a)Maicas, Carneiro and
  Bradley}]{maicas2017globally}
\bibinfo{author}{Maicas, G.}, \bibinfo{author}{Carneiro, G.},
  \bibinfo{author}{Bradley, A.P.}, \bibinfo{year}{2017}a.
\newblock \bibinfo{title}{Globally optimal breast mass segmentation from
  dce-mri using deep semantic segmentation as shape prior}, in:
  \bibinfo{booktitle}{2017 IEEE 14th International Symposium on Biomedical
  Imaging (ISBI 2017)}, pp. \bibinfo{pages}{305--309}.
\newblock \DOIprefix\doi{10.1109/ISBI.2017.7950525}.
%Type = Inproceedings
\bibitem[{Maicas et~al.(2017b)Maicas, Carneiro, Bradley, Nascimento and
  Reid}]{maicas2017deep}
\bibinfo{author}{Maicas, G.}, \bibinfo{author}{Carneiro, G.},
  \bibinfo{author}{Bradley, A.P.}, \bibinfo{author}{Nascimento, J.C.},
  \bibinfo{author}{Reid, I.}, \bibinfo{year}{2017}b.
\newblock \bibinfo{title}{Deep reinforcement learning for active breast lesion
  detection from dce-mri}, in: \bibinfo{booktitle}{Medical Image Computing and
  Computer-Assisted Intervention − MICCAI 2017}, \bibinfo{publisher}{Springer
  International Publishing}, \bibinfo{address}{Cham}. pp.
  \bibinfo{pages}{665--673}.
%Type = Article
\bibitem[{Maicas et~al.(2018b)Maicas, Snaauw, Bradley, Reid and
  Carneiro}]{maicas2018lesion}
\bibinfo{author}{Maicas, G.}, \bibinfo{author}{Snaauw, G.},
  \bibinfo{author}{Bradley, A.P.}, \bibinfo{author}{Reid, I.},
  \bibinfo{author}{Carneiro, G.}, \bibinfo{year}{2018}b.
\newblock \bibinfo{title}{Lesion saliency for weakly supervised lesion
  detection from breast dce-mri}.
\newblock \bibinfo{journal}{arXiv preprint arXiv:1807.07784} .
%Type = Article
\bibitem[{Mainiero et~al.(2017)Mainiero, Moy, Baron, Didwania, Green, Heller,
  Holbrook, Lee, Lewin, Lourenco et~al.}]{mainiero2017acr}
\bibinfo{author}{Mainiero, M.B.}, \bibinfo{author}{Moy, L.},
  \bibinfo{author}{Baron, P.}, \bibinfo{author}{Didwania, A.D.},
  \bibinfo{author}{Green, E.D.}, \bibinfo{author}{Heller, S.L.},
  \bibinfo{author}{Holbrook, A.I.}, \bibinfo{author}{Lee, S.J.},
  \bibinfo{author}{Lewin, A.A.}, \bibinfo{author}{Lourenco, A.P.}, et~al.,
  \bibinfo{year}{2017}.
\newblock \bibinfo{title}{Acr appropriateness criteria{\textregistered} breast
  cancer screening}.
\newblock \bibinfo{journal}{Journal of the American College of Radiology}
  \bibinfo{volume}{14}, \bibinfo{pages}{S383--S390}.
%Type = Article
\bibitem[{Mango et~al.(2015)Mango, Morris, Dershaw, Abramson, Fry, Moskowitz,
  Hughes, Kaplan and Jochelson}]{mango2015abbreviated}
\bibinfo{author}{Mango, V.L.}, \bibinfo{author}{Morris, E.A.},
  \bibinfo{author}{Dershaw, D.D.}, \bibinfo{author}{Abramson, A.},
  \bibinfo{author}{Fry, C.}, \bibinfo{author}{Moskowitz, C.S.},
  \bibinfo{author}{Hughes, M.}, \bibinfo{author}{Kaplan, J.},
  \bibinfo{author}{Jochelson, M.S.}, \bibinfo{year}{2015}.
\newblock \bibinfo{title}{Abbreviated protocol for breast mri: are multiple
  sequences needed for cancer detection?}
\newblock \bibinfo{journal}{European journal of radiology}
  \bibinfo{volume}{84}, \bibinfo{pages}{65--70}.
%Type = Article
\bibitem[{Matiisen et~al.(2017)Matiisen, Oliver, Cohen and
  Schulman}]{matiisen2017teacher}
\bibinfo{author}{Matiisen, T.}, \bibinfo{author}{Oliver, A.},
  \bibinfo{author}{Cohen, T.}, \bibinfo{author}{Schulman, J.},
  \bibinfo{year}{2017}.
\newblock \bibinfo{title}{Teacher-student curriculum learning}.
\newblock \bibinfo{journal}{arXiv preprint arXiv:1707.00183} .
%Type = Article
\bibitem[{Mcclymont(2015)}]{mcclymont2015computer}
\bibinfo{author}{Mcclymont, D.}, \bibinfo{year}{2015}.
\newblock \bibinfo{title}{Computer assisted detection and characterisation of
  breast cancer in mri} .
%Type = Article
\bibitem[{McClymont et~al.(2014)McClymont, Mehnert, Trakic
  et~al.}]{mcclymont2014fully}
\bibinfo{author}{McClymont, D.}, \bibinfo{author}{Mehnert, A.},
  \bibinfo{author}{Trakic, A.}, et~al., \bibinfo{year}{2014}.
\newblock \bibinfo{title}{Fully automatic lesion segmentation in breast {MRI}
  using mean-shift and graph-cuts on a region adjacency graph}.
\newblock \bibinfo{journal}{JMRI} \bibinfo{volume}{39},
  \bibinfo{pages}{795--804}.
\newblock \URLprefix \url{http://dx.doi.org/10.1002/jmri.24229},
  \DOIprefix\doi{10.1002/jmri.24229}.
%Type = Article
\bibitem[{Meinel et~al.(2007)Meinel, Stolpen, Berbaum, Fajardo and
  Reinhardt}]{meinel2007breast}
\bibinfo{author}{Meinel, L.A.}, \bibinfo{author}{Stolpen, A.H.},
  \bibinfo{author}{Berbaum, K.S.}, \bibinfo{author}{Fajardo, L.L.},
  \bibinfo{author}{Reinhardt, J.M.}, \bibinfo{year}{2007}.
\newblock \bibinfo{title}{Breast mri lesion classification: Improved
  performance of human readers with a backpropagation neural network
  computer-aided diagnosis (cad) system}.
\newblock \bibinfo{journal}{Journal of magnetic resonance imaging}
  \bibinfo{volume}{25}, \bibinfo{pages}{89--95}.
%Type = Article
\bibitem[{Milenkovi{\'c} et~al.(2017)Milenkovi{\'c}, Dalm{\i}{\c{s}},
  {\v{Z}}gajnar and Platel}]{milenkovic2017textural}
\bibinfo{author}{Milenkovi{\'c}, J.}, \bibinfo{author}{Dalm{\i}{\c{s}}, M.U.},
  \bibinfo{author}{{\v{Z}}gajnar, J.}, \bibinfo{author}{Platel, B.},
  \bibinfo{year}{2017}.
\newblock \bibinfo{title}{Textural analysis of early-phase spatiotemporal
  changes in contrast enhancement of breast lesions imaged with an ultrafast
  dce-mri protocol}.
\newblock \bibinfo{journal}{Medical physics} .
%Type = Article
\bibitem[{Mnih et~al.(2015)Mnih, Kavukcuoglu, Silver, Rusu, Veness, Bellemare,
  Graves, Riedmiller, Fidjeland, Ostrovski et~al.}]{mnih2015human}
\bibinfo{author}{Mnih, V.}, \bibinfo{author}{Kavukcuoglu, K.},
  \bibinfo{author}{Silver, D.}, \bibinfo{author}{Rusu, A.A.},
  \bibinfo{author}{Veness, J.}, \bibinfo{author}{Bellemare, M.G.},
  \bibinfo{author}{Graves, A.}, \bibinfo{author}{Riedmiller, M.},
  \bibinfo{author}{Fidjeland, A.K.}, \bibinfo{author}{Ostrovski, G.}, et~al.,
  \bibinfo{year}{2015}.
\newblock \bibinfo{title}{Human-level control through deep reinforcement
  learning}.
\newblock \bibinfo{journal}{Nature} \bibinfo{volume}{518},
  \bibinfo{pages}{529}.
%Type = Article
\bibitem[{Mus et~al.(2017)Mus, Borelli, Bult, Weiland, Karssemeijer, Barentsz,
  Gubern-M{\'e}rida, Platel and Mann}]{mus2017time}
\bibinfo{author}{Mus, R.D.}, \bibinfo{author}{Borelli, C.},
  \bibinfo{author}{Bult, P.}, \bibinfo{author}{Weiland, E.},
  \bibinfo{author}{Karssemeijer, N.}, \bibinfo{author}{Barentsz, J.O.},
  \bibinfo{author}{Gubern-M{\'e}rida, A.}, \bibinfo{author}{Platel, B.},
  \bibinfo{author}{Mann, R.M.}, \bibinfo{year}{2017}.
\newblock \bibinfo{title}{Time to enhancement derived from ultrafast breast mri
  as a novel parameter to discriminate benign from malignant breast lesions}.
\newblock \bibinfo{journal}{European journal of radiology}
  \bibinfo{volume}{89}, \bibinfo{pages}{90--96}.
%Type = Article
\bibitem[{Platel et~al.(2014)Platel, Mus, Welte, Karssemeijer and
  Mann}]{platel2014automated}
\bibinfo{author}{Platel, B.}, \bibinfo{author}{Mus, R.},
  \bibinfo{author}{Welte, T.}, \bibinfo{author}{Karssemeijer, N.},
  \bibinfo{author}{Mann, R.}, \bibinfo{year}{2014}.
\newblock \bibinfo{title}{Automated characterization of breast lesions imaged
  with an ultrafast dce-mr protocol}.
\newblock \bibinfo{journal}{IEEE transactions on medical imaging}
  \bibinfo{volume}{33}, \bibinfo{pages}{225--232}.
%Type = Article
\bibitem[{Rasti et~al.(2017)Rasti, Teshnehlab and Phung}]{rasti2017breast}
\bibinfo{author}{Rasti, R.}, \bibinfo{author}{Teshnehlab, M.},
  \bibinfo{author}{Phung, S.L.}, \bibinfo{year}{2017}.
\newblock \bibinfo{title}{Breast cancer diagnosis in dce-mri using mixture
  ensemble of convolutional neural networks}.
\newblock \bibinfo{journal}{Pattern Recognition} \bibinfo{volume}{72},
  \bibinfo{pages}{381--390}.
%Type = Inproceedings
\bibitem[{Ren et~al.(2015)Ren, He, Girshick and Sun}]{ren2015faster}
\bibinfo{author}{Ren, S.}, \bibinfo{author}{He, K.}, \bibinfo{author}{Girshick,
  R.}, \bibinfo{author}{Sun, J.}, \bibinfo{year}{2015}.
\newblock \bibinfo{title}{Faster r-cnn: Towards real-time object detection with
  region proposal networks}, in: \bibinfo{booktitle}{Advances in neural
  information processing systems}, pp. \bibinfo{pages}{91--99}.
%Type = Article
\bibitem[{Renz et~al.(2012)Renz, B{\"o}ttcher, Diekmann, Poellinger, Maurer,
  Pfeil, Streitparth, Collettini, Bick, Hamm et~al.}]{renz2012detection}
\bibinfo{author}{Renz, D.M.}, \bibinfo{author}{B{\"o}ttcher, J.},
  \bibinfo{author}{Diekmann, F.}, \bibinfo{author}{Poellinger, A.},
  \bibinfo{author}{Maurer, M.H.}, \bibinfo{author}{Pfeil, A.},
  \bibinfo{author}{Streitparth, F.}, \bibinfo{author}{Collettini, F.},
  \bibinfo{author}{Bick, U.}, \bibinfo{author}{Hamm, B.}, et~al.,
  \bibinfo{year}{2012}.
\newblock \bibinfo{title}{Detection and classification of contrast-enhancing
  masses by a fully automatic computer-assisted diagnosis system for breast
  mri}.
\newblock \bibinfo{journal}{Journal of Magnetic Resonance Imaging}
  \bibinfo{volume}{35}, \bibinfo{pages}{1077--1088}.
%Type = Article
\bibitem[{Ribli et~al.(2018)Ribli, Horv{\'a}th, Unger, Pollner and
  Csabai}]{ribli2018detecting}
\bibinfo{author}{Ribli, D.}, \bibinfo{author}{Horv{\'a}th, A.},
  \bibinfo{author}{Unger, Z.}, \bibinfo{author}{Pollner, P.},
  \bibinfo{author}{Csabai, I.}, \bibinfo{year}{2018}.
\newblock \bibinfo{title}{Detecting and classifying lesions in mammograms with
  deep learning}.
\newblock \bibinfo{journal}{Scientific reports} \bibinfo{volume}{8},
  \bibinfo{pages}{4165}.
%Type = Inproceedings
\bibitem[{Ronneberger et~al.(2015)Ronneberger, Fischer and
  Brox}]{ronneberger2015u}
\bibinfo{author}{Ronneberger, O.}, \bibinfo{author}{Fischer, P.},
  \bibinfo{author}{Brox, T.}, \bibinfo{year}{2015}.
\newblock \bibinfo{title}{U-net: Convolutional networks for biomedical image
  segmentation}, in: \bibinfo{booktitle}{International Conference on Medical
  image computing and computer-assisted intervention},
  \bibinfo{organization}{Springer}. pp. \bibinfo{pages}{234--241}.
%Type = Article
\bibitem[{Saadatmand et~al.(2015)Saadatmand, Bretveld, Siesling and
  Tilanus-Linthorst}]{saadatmand2015influence}
\bibinfo{author}{Saadatmand, S.}, \bibinfo{author}{Bretveld, R.},
  \bibinfo{author}{Siesling, S.}, \bibinfo{author}{Tilanus-Linthorst, M.M.},
  \bibinfo{year}{2015}.
\newblock \bibinfo{title}{Influence of tumour stage at breast cancer detection
  on survival in modern times: population based study in 173 797 patients}.
\newblock \bibinfo{journal}{Bmj} \bibinfo{volume}{351}, \bibinfo{pages}{h4901}.
%Type = Inproceedings
\bibitem[{Selvaraju et~al.(2017)Selvaraju, Cogswell, Das, Vedantam, Parikh and
  Batra}]{selvaraju2017grad}
\bibinfo{author}{Selvaraju, R.R.}, \bibinfo{author}{Cogswell, M.},
  \bibinfo{author}{Das, A.}, \bibinfo{author}{Vedantam, R.},
  \bibinfo{author}{Parikh, D.}, \bibinfo{author}{Batra, D.},
  \bibinfo{year}{2017}.
\newblock \bibinfo{title}{Grad-cam: Visual explanations from deep networks via
  gradient-based localization}, in: \bibinfo{booktitle}{Computer Vision (ICCV),
  2017 IEEE International Conference on}, \bibinfo{organization}{IEEE}. pp.
  \bibinfo{pages}{618--626}.
%Type = Article
\bibitem[{Shimauchi et~al.(2011)Shimauchi, Giger, Bhooshan, Lan, Pesce, Lee,
  Abe and Newstead}]{shimauchi2011evaluation}
\bibinfo{author}{Shimauchi, A.}, \bibinfo{author}{Giger, M.L.},
  \bibinfo{author}{Bhooshan, N.}, \bibinfo{author}{Lan, L.},
  \bibinfo{author}{Pesce, L.L.}, \bibinfo{author}{Lee, J.K.},
  \bibinfo{author}{Abe, H.}, \bibinfo{author}{Newstead, G.M.},
  \bibinfo{year}{2011}.
\newblock \bibinfo{title}{Evaluation of clinical breast mr imaging performed
  with prototype computer-aided diagnosis breast mr imaging workstation: reader
  study}.
\newblock \bibinfo{journal}{Radiology} \bibinfo{volume}{258},
  \bibinfo{pages}{696--704}.
%Type = Article
\bibitem[{Siegel et~al.(2017)Siegel, Miller and Jemal}]{cancerStats17}
\bibinfo{author}{Siegel, R.L.}, \bibinfo{author}{Miller, K.D.},
  \bibinfo{author}{Jemal, A.}, \bibinfo{year}{2017}.
\newblock \bibinfo{title}{Cancer statistics, 2017}.
\newblock \bibinfo{journal}{CA: A Cancer Journal for Clinicians} .
%Type = Article
\bibitem[{Smith et~al.(2017)Smith, Andrews, Brooks, Fedewa,
  Manassaram-Baptiste, Saslow, Brawley and Wender}]{smith2017cancer}
\bibinfo{author}{Smith, R.A.}, \bibinfo{author}{Andrews, K.S.},
  \bibinfo{author}{Brooks, D.}, \bibinfo{author}{Fedewa, S.A.},
  \bibinfo{author}{Manassaram-Baptiste, D.}, \bibinfo{author}{Saslow, D.},
  \bibinfo{author}{Brawley, O.W.}, \bibinfo{author}{Wender, R.C.},
  \bibinfo{year}{2017}.
\newblock \bibinfo{title}{Cancer screening in the united states, 2017: a review
  of current american cancer society guidelines and current issues in cancer
  screening}.
\newblock \bibinfo{journal}{CA: a cancer journal for clinicians}
  \bibinfo{volume}{67}, \bibinfo{pages}{100--121}.
%Type = Article
\bibitem[{Soares et~al.(2013)Soares, Janela, Pereira, Seabra and
  Freire}]{soares20133d}
\bibinfo{author}{Soares, F.}, \bibinfo{author}{Janela, F.},
  \bibinfo{author}{Pereira, M.}, \bibinfo{author}{Seabra, J.},
  \bibinfo{author}{Freire, M.M.}, \bibinfo{year}{2013}.
\newblock \bibinfo{title}{3d lacunarity in multifractal analysis of breast
  tumor lesions in dynamic contrast-enhanced magnetic resonance imaging}.
\newblock \bibinfo{journal}{IEEE Transactions on Image Processing}
  \bibinfo{volume}{22}, \bibinfo{pages}{4422--4435}.
%Type = Article
\bibitem[{Song et~al.(2016)Song, Chen, Yuan and Sun}]{song2016progress}
\bibinfo{author}{Song, J.L.}, \bibinfo{author}{Chen, C.},
  \bibinfo{author}{Yuan, J.P.}, \bibinfo{author}{Sun, S.R.},
  \bibinfo{year}{2016}.
\newblock \bibinfo{title}{Progress in the clinical detection of heterogeneity
  in breast cancer}.
\newblock \bibinfo{journal}{Cancer medicine} \bibinfo{volume}{5},
  \bibinfo{pages}{3475--3488}.
%Type = Book
\bibitem[{Sutton and Barto(1998)}]{Sut1998}
\bibinfo{author}{Sutton, R.}, \bibinfo{author}{Barto, A.G.},
  \bibinfo{year}{1998}.
\newblock \bibinfo{title}{{Reinforcement learning: An introduction}}.
  volume~\bibinfo{volume}{2}.
\newblock \bibinfo{publisher}{MIT press}.
%Type = Article
\bibitem[{Tajbakhsh et~al.(2016)Tajbakhsh, Shin, Gurudu, Hurst, Kendall, Gotway
  and Liang}]{tajbakhsh2016convolutional}
\bibinfo{author}{Tajbakhsh, N.}, \bibinfo{author}{Shin, J.Y.},
  \bibinfo{author}{Gurudu, S.R.}, \bibinfo{author}{Hurst, R.T.},
  \bibinfo{author}{Kendall, C.B.}, \bibinfo{author}{Gotway, M.B.},
  \bibinfo{author}{Liang, J.}, \bibinfo{year}{2016}.
\newblock \bibinfo{title}{Convolutional neural networks for medical image
  analysis: Full training or fine tuning?}
\newblock \bibinfo{journal}{IEEE transactions on medical imaging}
  \bibinfo{volume}{35}, \bibinfo{pages}{1299--1312}.
%Type = Article
\bibitem[{Torre et~al.(2015)Torre, Bray, Siegel, Ferlay, Lortet-Tieulent and
  Jemal}]{torre2015global}
\bibinfo{author}{Torre, L.A.}, \bibinfo{author}{Bray, F.},
  \bibinfo{author}{Siegel, R.L.}, \bibinfo{author}{Ferlay, J.},
  \bibinfo{author}{Lortet-Tieulent, J.}, \bibinfo{author}{Jemal, A.},
  \bibinfo{year}{2015}.
\newblock \bibinfo{title}{Global cancer statistics, 2012}.
\newblock \bibinfo{journal}{CA: a cancer journal for clinicians}
  \bibinfo{volume}{65}, \bibinfo{pages}{87--108}.
%Type = Article
\bibitem[{Vreemann et~al.(2018)Vreemann, Gubern-Merida, Lardenoije, Bult,
  Karssemeijer, Pinker and Mann}]{Vreemann2018}
\bibinfo{author}{Vreemann, S.}, \bibinfo{author}{Gubern-Merida, A.},
  \bibinfo{author}{Lardenoije, S.}, \bibinfo{author}{Bult, P.},
  \bibinfo{author}{Karssemeijer, N.}, \bibinfo{author}{Pinker, K.},
  \bibinfo{author}{Mann, R.M.}, \bibinfo{year}{2018}.
\newblock \bibinfo{title}{The frequency of missed breast cancers in women
  participating in a high-risk mri screening program}.
\newblock \bibinfo{journal}{Breast Cancer Research and Treatment} \URLprefix
  \url{https://doi.org/10.1007/s10549-018-4688-z},
  \DOIprefix\doi{10.1007/s10549-018-4688-z}.
%Type = Inproceedings
\bibitem[{Wang et~al.(2014)Wang, Harz, Boehler, Platel, Homeyer and
  Hahn}]{wang2014robust}
\bibinfo{author}{Wang, L.}, \bibinfo{author}{Harz, M.},
  \bibinfo{author}{Boehler, T.}, \bibinfo{author}{Platel, B.},
  \bibinfo{author}{Homeyer, A.}, \bibinfo{author}{Hahn, H.K.},
  \bibinfo{year}{2014}.
\newblock \bibinfo{title}{A robust and extendable framework towards fully
  automated diagnosis of nonmass lesions in breast dce-mri}, in:
  \bibinfo{booktitle}{International Symposium on Biomedical Imaging},
  \bibinfo{organization}{IEEE}. pp. \bibinfo{pages}{129--132}.
%Type = Inproceedings
\bibitem[{Wang et~al.(2017a)Wang, Peng, Lu, Lu, Bagheri and
  Summers}]{wang2017chestx}
\bibinfo{author}{Wang, X.}, \bibinfo{author}{Peng, Y.}, \bibinfo{author}{Lu,
  L.}, \bibinfo{author}{Lu, Z.}, \bibinfo{author}{Bagheri, M.},
  \bibinfo{author}{Summers, R.M.}, \bibinfo{year}{2017}a.
\newblock \bibinfo{title}{Chestx-ray8: Hospital-scale chest x-ray database and
  benchmarks on weakly-supervised classification and localization of common
  thorax diseases}, in: \bibinfo{booktitle}{2017 IEEE Conference on Computer
  Vision and Pattern Recognition (CVPR)}.
%Type = Inproceedings
\bibitem[{Wang et~al.(2017b)Wang, Yin, Shi, Fang, Li and Wang}]{wang2017zoom}
\bibinfo{author}{Wang, Z.}, \bibinfo{author}{Yin, Y.}, \bibinfo{author}{Shi,
  J.}, \bibinfo{author}{Fang, W.}, \bibinfo{author}{Li, H.},
  \bibinfo{author}{Wang, X.}, \bibinfo{year}{2017}b.
\newblock \bibinfo{title}{Zoom-in-net: Deep mining lesions for diabetic
  retinopathy detection}, in: \bibinfo{booktitle}{International Conference on
  Medical Image Computing and Computer-Assisted Intervention},
  \bibinfo{organization}{Springer}. pp. \bibinfo{pages}{267--275}.
%Type = Article
\bibitem[{Welch et~al.(2016)Welch, Prorok, O’Malley and
  Kramer}]{welch2016breast}
\bibinfo{author}{Welch, H.G.}, \bibinfo{author}{Prorok, P.C.},
  \bibinfo{author}{O’Malley, A.J.}, \bibinfo{author}{Kramer, B.S.},
  \bibinfo{year}{2016}.
\newblock \bibinfo{title}{Breast-cancer tumor size, overdiagnosis, and
  mammography screening effectiveness}.
\newblock \bibinfo{journal}{New England Journal of Medicine}
  \bibinfo{volume}{375}, \bibinfo{pages}{1438--1447}.
%Type = Article
\bibitem[{Wood(2005)}]{wood2005computer}
\bibinfo{author}{Wood, C.}, \bibinfo{year}{2005}.
\newblock \bibinfo{title}{Computer aided detection (cad) for breast mri}.
\newblock \bibinfo{journal}{Technology in cancer research \& treatment}
  \bibinfo{volume}{4}, \bibinfo{pages}{49--53}.
%Type = Article
\bibitem[{Xue et~al.(2018)Xue, Brahm, Pandey, Leung and Li}]{xue2018full}
\bibinfo{author}{Xue, W.}, \bibinfo{author}{Brahm, G.},
  \bibinfo{author}{Pandey, S.}, \bibinfo{author}{Leung, S.},
  \bibinfo{author}{Li, S.}, \bibinfo{year}{2018}.
\newblock \bibinfo{title}{Full left ventricle quantification via deep multitask
  relationships learning}.
\newblock \bibinfo{journal}{Medical image analysis} \bibinfo{volume}{43},
  \bibinfo{pages}{54--65}.
%Type = Inproceedings
\bibitem[{Yang et~al.(2017)Yang, Wang, Liu, Le, Chen, Cheng and
  Wang}]{yang2017joint}
\bibinfo{author}{Yang, X.}, \bibinfo{author}{Wang, Z.}, \bibinfo{author}{Liu,
  C.}, \bibinfo{author}{Le, H.M.}, \bibinfo{author}{Chen, J.},
  \bibinfo{author}{Cheng, K.T.T.}, \bibinfo{author}{Wang, L.},
  \bibinfo{year}{2017}.
\newblock \bibinfo{title}{Joint detection and diagnosis of prostate cancer in
  multi-parametric mri based on multimodal convolutional neural networks}, in:
  \bibinfo{booktitle}{International Conference on Medical Image Computing and
  Computer-Assisted Intervention}, \bibinfo{organization}{Springer}. pp.
  \bibinfo{pages}{426--434}.
%Type = Inproceedings
\bibitem[{Zeiler and Fergus(2014)}]{zeiler2014visualizing}
\bibinfo{author}{Zeiler, M.D.}, \bibinfo{author}{Fergus, R.},
  \bibinfo{year}{2014}.
\newblock \bibinfo{title}{Visualizing and understanding convolutional
  networks}, in: \bibinfo{booktitle}{ECCV}.
%Type = Inproceedings
\bibitem[{Zhou et~al.(2016)Zhou, Khosla, Lapedriza, Oliva and
  Torralba}]{zhou2016learning}
\bibinfo{author}{Zhou, B.}, \bibinfo{author}{Khosla, A.},
  \bibinfo{author}{Lapedriza, A.}, \bibinfo{author}{Oliva, A.},
  \bibinfo{author}{Torralba, A.}, \bibinfo{year}{2016}.
\newblock \bibinfo{title}{Learning deep features for discriminative
  localization}, in: \bibinfo{booktitle}{Proceedings of the IEEE Conference on
  Computer Vision and Pattern Recognition}, pp. \bibinfo{pages}{2921--2929}.
%Type = Inproceedings
\bibitem[{Zhu et~al.(2017)Zhu, Lou, Vang and Xie}]{zhu2017deep}
\bibinfo{author}{Zhu, W.}, \bibinfo{author}{Lou, Q.}, \bibinfo{author}{Vang,
  Y.S.}, \bibinfo{author}{Xie, X.}, \bibinfo{year}{2017}.
\newblock \bibinfo{title}{Deep multi-instance networks with sparse label
  assignment for whole mammogram classification}, in:
  \bibinfo{booktitle}{International Conference on Medical Image Computing and
  Computer-Assisted Intervention}.

\end{thebibliography}






% \end{thebibliography}

\end{document}